\documentclass[10pt,twocolumn,letterpaper]{article}

\usepackage{iccv} 
\usepackage{times}
\usepackage{epsfig}
\usepackage{graphicx}
\usepackage{amsmath}
\usepackage{amssymb}
\usepackage{bm}
\usepackage{caption}
\usepackage{multirow}

\iccvfinalcopy

\usepackage{booktabs}

\usepackage{dsfont}

\usepackage{color}
\usepackage{xcolor}

\usepackage{algorithm}
\usepackage[noend]{algpseudocode}

\usepackage{rotating}
\usepackage{tabularx}
\usepackage{threeparttable}

\usepackage{tcolorbox}
\usepackage{soul}

\usepackage[capitalize]{cleveref}
\crefname{section}{Sec.}{Secs.}
\Crefname{section}{Section}{Sections}
\Crefname{table}{Table}{Tables}
\crefname{table}{Tab.}{Tabs.}

\newcommand{\mycaption}[2]{\caption{\textbf{#1.}\xspace#2}}

\begin{document}

\title{Near Perfect GAN Inversion}

\author{
Qianli Feng\\
Amazon\\
\and
Viraj Shah\\
Amazon\\
\and
Raghudeep Gadde\\
Amazon\\
\and
Pietro Perona\\
Amazon\\
\and
Aleix Martinez\\
Amazon\\
}
\twocolumn[{\renewcommand\twocolumn[1][]{#1}%
\maketitle
\vspace{-1.1cm}
\begin{center}
    \includegraphics[width=\textwidth]{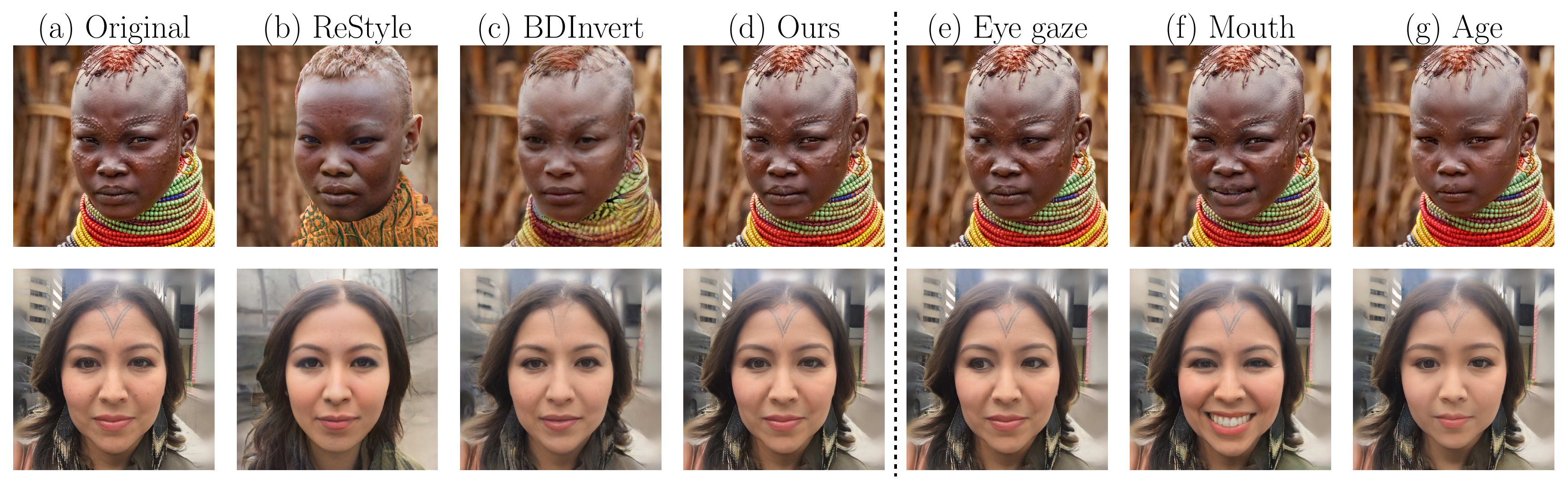}
\end{center}
\vspace{-.5cm}
\captionof{figure}{(a) Two real photos. The goal of GAN inversion is to find a latent code that synthesizes an image that is as similar as possible to that in (a). Synthesizing these photos is hard as made evident by the results obtained using state-of-the-art GAN inversion algorithms as shown in (b) and (c). In (d) we show the results of the method proposed in this paper. As we can see, the synthesized images are almost identical clones with only tiny differences not visible to the naked eye. Furthermore, these clones are readily editable using standard GAN editing algorithms as shown in (e-f). Results best seen at a 600\% zoom.}
\label{fig:teaser}
\vspace{.25cm}
}]

\begin{abstract}

To edit a real photo using Generative Adversarial Networks (GANs), we need a GAN inversion algorithm to identify the latent vector that perfectly reproduces it. Unfortunately, whereas existing inversion algorithms can synthesize images similar to real photos, they cannot generate the identical clones needed in most applications. Here, we derive an algorithm that achieves near perfect reconstructions of photos. Rather than relying on encoder- or optimization-based methods to find an inverse mapping on a fixed generator $G(\cdot)$, we derive an approach to locally adjust $G(\cdot)$ to more optimally represent the photos we wish to synthesize. This is done by locally tweaking the learned mapping $G(\cdot)$ s.t. $\| {\bf x} - G({\bf z}) \|<\epsilon$, with ${\bf x}$ the photo we wish to reproduce, ${\bf z}$ the latent vector, $\|\cdot\|$ an appropriate metric, and $\epsilon > 0$ a small scalar.  We show that this approach can not only produce synthetic images that are indistinguishable from the real photos we wish to replicate, but that these images are readily editable. We demonstrate the effectiveness of the derived algorithm on a variety of datasets including human faces, animals, and cars, and discuss its importance for diversity and inclusion.

\end{abstract}

\section{Introduction}
\label{sec:introduction}

Generative Adversarial Networks (GANs) have seen dramatic improvement on image photo-realism in recent years \cite{Gadde_2021_ICCV,Karras2019stylegan2,Karras2021,brock2018large,karras2019stylebased}. Crucially, GAN-generated images are editable, enabling a number of previously difficult to envision applications. For example, in movies, ads, virtual reality, video games and e-commerce, we can now edit existing photos to improve or showcase image variants that are difficult, costly, or impossible to film. For instance, a face can be shown to be younger or have a different hair color or style.

To edit real photos though, we need a GAN inversion algorithm, \cref{fig:teaser}. More formally, the generator function $G({\bf z})={\bf x}$ maps a latent vector ${\bf z}\in\mathcal{Z}\subseteq \mathbb{R}^q$ into an image ${\bf x}\in\mathcal{X}\subseteq \mathbb{R}^p$, with $q \ll p$. GAN inversion is the problem of finding ${\bf z}$ given ${\bf x}$, i.e., ${\bf z}=G^{-1}({\bf x})$. However, unlike Normalizing Flows and auto-encoders, inverting $G(\cdot)$ in GANs is generally impossible. To solve this problem, researchers use encoder or optimization approaches to solve
\begin{equation}
\label{eq:gan-inversion}
	\min \| G(h(\mathbf{x})) - \mathbf{x} \| 
\end{equation}
with the algorithm converging to a solution iff $\| G(h(\mathbf{x})) - \mathbf{x} \| < \epsilon$, and where $\| \cdot \|$ is a metric in image space $\mathcal{X}$, $h: \mathcal{X} \mapsto \mathcal{Z}$, and $\epsilon > 0$ is small. 

Optimization algorithms find the latent vector representation of a given image by optimizing $h(\cdot)$ as in \cref{eq:gan-inversion} from a starting guess \cite{Lipton2017PreciseRO,Creswell2019InvertingTG,Abdal2019Image2StyleGANHT,Abdal2020Image2StyleGANHT}. In encoder approaches $h(\cdot)$ is usually a deep network trained to map from $\mathcal{X}$ onto $\mathcal{Z}$ and optimize \cref{eq:gan-inversion} over a training set \cite{pidhorskyi2020adversarial,pidhorskyi2020adversarial,Richardson2020EncodingIS,alaluf2021restyle,Chai2021UsingLS}. There are also hybrid approaches \cite{Chai2021EnsemblingWD,Bau2019SeeingWA}. 

Hence, crucially, optimization and encoder approaches assume that the mapping $G(\cdot)$ is able to synthesize the desirable photo ${\bf x}$. However, given the large variety of all possible photos $\mathcal{X}$, this assumption will typically not hold, resulting in sub-par reconstruction results, \cref{fig:teaser}(b, c). 

While optimization- and encoder-based approaches optimize or learn $h(\cdot)$ while fixing $G(\cdot)$, our key contribution is to show we can fix $h(\cdot)$ and locally tune $G(\cdot)$ instead. 

\cref{fig:teaser}(d) shows the reconstruction results of our approach compared to state-of-the-art methods \cref{fig:teaser}(b,c). GAN editing can then be applied to the reconstructed images, yielding photo-realistic image variants, \cref{fig:teaser}(e-g).

The derived solution is not only scientifically novel, it is also essential for many real-world applications. Semantic manipulation of real photos is only meaningful if the edited images are seen as realistic variants of the original photos, \eg, the exact same face and background but with a different eye gaze, mouth shape or facial hair as illustrated in \cref{fig:teaser}(e-g). Additionally, as out-of-sample photos are more likely to be ill-inverted, most current algorithms cannot be used on photos of groups under-represented in the training data, which may include ethnic, racial, gender, age, religious, cultural, job/occupation, etc. Our solution solves these important limitations. 

We provide extensive comparative evaluations on several datasets against the state of the art, showing that our reconstructed images keep their photo-realism even after being edited, and demonstrate that our derived algorithm is equally applicable to under-represented groups, increasing diversity and inclusion.

\section{Related Work}
\label{sec:relatedworks}
 
Two GAN inversion algorithms have been proposed in the literature: ``projection via optimization'' and ``projection via a feedforward network'', \ie, optimization- and an encoder-based method \cite{Zhu2016GenerativeVM}. 
 
\subsection{Optimization-based inversion}
 
Optimization-based methods require making three choices. First, a loss to measure the similarity between the real photo and the reconstructed image; second,  a latent space over which the loss function may be optimized; and, third, the optimization criterion. Early works like \cite{Lipton2017PreciseRO} explore the usage of the likelihood loss in a Gaussian latent space. Others \cite{Creswell2019InvertingTG} perform stochastic clipping, limiting the magnitude of the gradient as it optimizes image similarity. 

\subsection{Encoder-based inversion}
 
For encoder-based or learning-based methods as in \cite{xia2021gan}, we require first, an encoder network; second, a loss to measure the similarity between the real photo and the reconstructed image, and third, the output vectors on the latent space. \cite{pidhorskyi2020adversarial} proposed an auto-encoder architecture incorporating a StyleGAN generator, with an explicit mapping learnt from synthetic image. Others use different mapping or latent space representations \cite{Richardson2020EncodingIS}.

Compared to optimization-based methods, encoder-based algorithms enjoy the benefit of faster inference time. To further improve their results, ReStyle \cite{alaluf2021restyle} defines an iterative residual-based encoder. \cite{Chai2021UsingLS} proposed a latent space encoder with masked input to study the compositionality in GANs latent space. There are also hybrid methods \cite{Chai2021EnsemblingWD,Bau2019SeeingWA}, combining the advantages of encoder- and optimization-based method.  
 
\subsection{Limitations of these methods}
 
The pre-trained generator is fixed in both the optimization- and encoder-based methods. Since there is no guarantee that the photos we wish to replicate synthetically can  be generated by $G(\cdot)$, these methods typically yield disappointing results (Fig.~\ref{fig:teaser}). Moreover, groups that are under-represented in the training set cannot be reproduced accurately, lowering diversity and decreasing inclusion.
 
In our work we propose a third way: we allow the generator to be updated locally about the query latent vector. We optimize this update to obtain a faithful reproduction of the photo of interest. This yields better reconstruction accuracy while maintaining editability of the image. 
 

\begin{figure}
  \centering
  \includegraphics[width=\columnwidth]{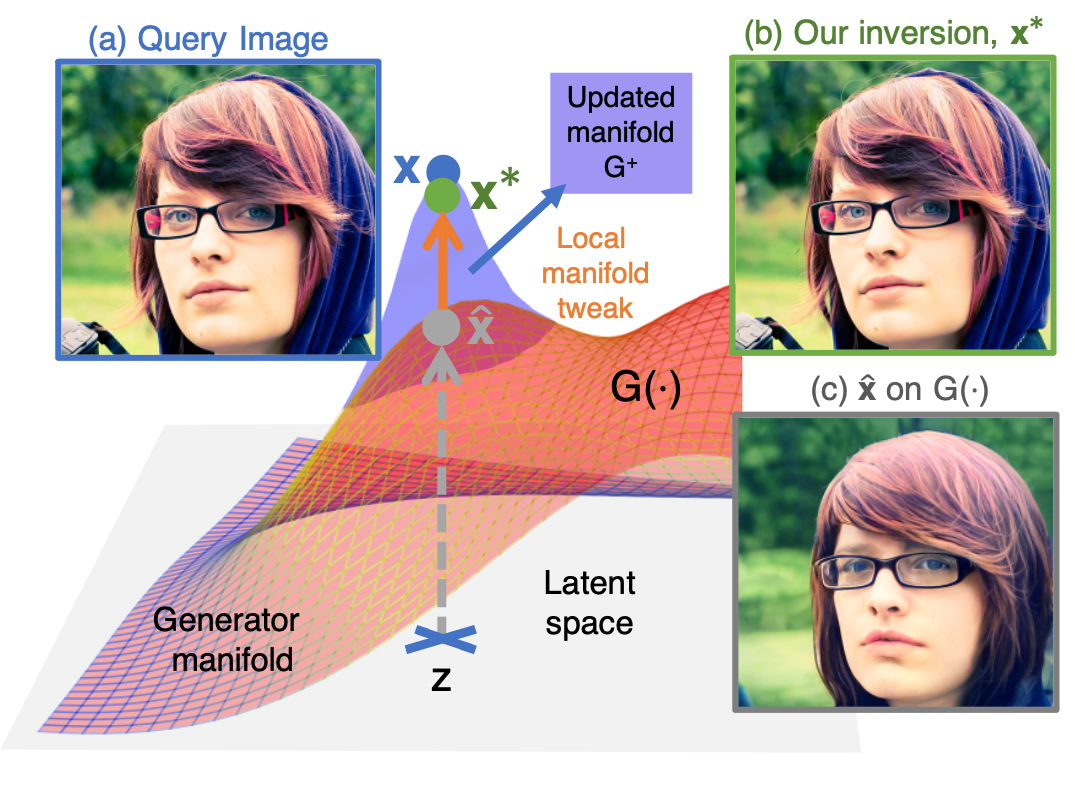}
  \caption{The query photo we wish to synthesize is given by ${\bf x}$. Given a pre-trained GAN model, its generator function $G(\cdot)$ can, at best, synthesize image $\hat{\bf x}$. That is, $\hat{\bf x}$ is the point on the manifold that is closest to ${\bf x}$. This paper proposes an algorithm to locally tweak $G(\cdot)$ to include an image ${\bf x}^*$ s.t. $\| {\bf x} - {\bf x}^* \| < \epsilon$. We refer to the new tweaked manifold as $G^+(\cdot)$. The original $G(\cdot)$ is show in orange and the tweaked $G^+(\cdot)$ in blue. The latent space is represented as a gray plane at the bottom.}
  \vspace{-0.5em}
  \label{fig:method-overview}
\end{figure}

\section{Method}
\label{sec:method}

\subsection{Problem definition}\label{subsec:problem-definition}
For a given query photo $\mathbf{x}\in \mathcal{X} $, we want to obtain its corresponding latent code $\mathbf{z}\in \mathcal{Z} $ that reconstructs $\mathbf{x}$ as accurately as possible, \ie, $G({\bf z})={\bf x}^*$, with $\| {\bf x}^* - {\bf x} \| < \epsilon$. 

Previous encoder- and optimization-based methods focus on optimizing $h(\cdot)$ while keeping $G(\cdot)$ frozen.  To reconstruct $\mathbf{x}$, these previous methods have to operate under the assumption that  ${\bf x}^*$ is on the manifold defined by $G(\cdot)$, \ie, $\exists \, {\bf z} \in \mathcal{Z} \text{ s.t. } G({\bf z})={\bf x}^*$. 

When $\mathbf{x}$ does not lie on or very close to the pre-trained manifold defined by $G(\cdot)$, the best these methods can do is to retrieve its nearest projection $\mathbf{\hat{x}}$ as illustrated in \cref{fig:method-overview}. This figure shows an example where the query photo ${\bf x}$ is not on the manifold defined by $G(\cdot)$, which is shown as an orange manifold. Thus, GAN inversion methods can at best synthesize the image $\hat{\bf x}$. $\hat{\bf x}$ is the image on the manifold defined by $G(\cdot)$ that is closest to ${\bf x}$; here closeness is given by an orthographic projection onto the manifold. Our proposal is an algorithm that locally tweaks this manifold $G(\cdot)$ to include the image ${\bf x}^*$, an image that is as close as possible to the query photo ${\bf x}$. This is shown as a blue extension of the manifold in the figure. 

\subsection{Tweaking the manifold locally}\label{subsec:hot-inversion}

Let us now derive the method to update the manifold defined by the generator locally. 

Note we cannot simply modify the manifold $G(\cdot)$ in any random way that happens to include ${\bf x}$. This is because in addition to including our query image ${\bf x}$, the manifold should only change locally and in a way that allows us to edit the synthesized version of ${\bf x}$ as easily as we edit any other synthetic image. In addition, we need to ensure that these edits yield synthetic image variants that look as realistic as the original photo.

To successfully edit the manifold locally, we first need to find $\hat{\bf x}$, \ie, the closest point to ${\bf x}$ we can find on the manifold. To this end, we can use any of the existing approaches described above. That is, we optimize $h(\cdot)$ by keeping $G(\cdot)$ fix.

Once we have $\hat{\bf x}$, we fix $h(\cdot)$ and let $G(\cdot)$ change locally about $\hat{\bf x}$ to include ${\bf x}^*$, \cref{fig:method-overview}. Our goal is to make the smallest change possible while maintaining the desirable properties of the pre-trained $G(\cdot)$, \eg, we can edit images in a number of controllable ways. 

We do this by combining two loss functions. The first loss function $\mathcal{L}_\text{local} $ is tasked to locally tweak the manifold to include ${\bf x}$ by making the distance from ${\bf x}$ to ${\bf x}^*$ as small as possible and keeping the properties of the manifold intact. The second loss function $\mathcal{L}_\text{global}$ ensures the rest of the manifold does not change.

\subsection{Local loss}

For the manifold defined by $G(\cdot)$ to generate ${\bf x}$, there needs to be a latent vector ${\bf z}$ s.t. $G({\bf z})$ is as similar to ${\bf x}$ as possible. We can compute this using a reconstruction loss function $\mathcal{L}_\text{recon} ({\bf x}_1,{\bf x}_2)$ that measure the similarity between ${\bf x}_1$ and ${\bf x}_2$, with ${\bf x}_i \in \mathcal{X}$. 

Because our goal is to synthesize an image that is as visually similar to the query photo as possible, we choose to use the Laplacian pyramid \cite{adelson1984pyramid,bojanowski2018optimizing} loss function as $\mathcal{L}_\text{recon} (\cdot,\cdot)$. Note, however, that other similarity losses could be used.

Let $\mathcal{L}_\text{recon} = \text{LaplacianPyramid}(\mathbf{x},G({\bf z}))$ be the reconstruction loss computed using the Laplacian pyramid, calculated by summing over mean L1 differences across all levels of a Laplacian pyramid of ${\bf x}$ and $G({\bf z})$. We can now find ${\bf z}$ s.t. $G({{\bf z}})=\hat{\bf x}$ and then optimize $\mathcal{L}_\text{recon}$ until $\mathcal{L}_\text{recon}<\epsilon$. 

To encourage that the tweaked manifold is editable and maintains all other desirable properties, we regularize this solution with an adversarial loss, $\mathcal{L}_\text{adv\_local}$. Specifically, we consider, $\mathcal{L}_\text{adv\_local} = \log D(\mathbf{x}) + \log (1-D(G({\bf z})))$, where $D(\cdot)$ is the discriminator.

The combined local loss is thus given by,
\begin{equation}\label{eq:local-loss}
	\mathcal{L}_\text{local} = \mathcal{L}_\text{recon} + \lambda \mathcal{L}_\text{adv\_local},
\end{equation}
where $\lambda$ is the regularizing term. Here too we find ${\bf z}$ s.t. $G({{\bf z}})=\hat{\bf x}$ and then optimize $\mathcal{L}_\text{local}$.

\subsection{Global cohesion}

We still need to ensure that the rest of the manifold does not change. This is to make sure that the model keeps any previous training and tweaking we have applied. We do this by computing a global loss function to enforce overall stability of the manifold.

To this end we use the loss of the pre-trained GAN model. For example, when using StyleGAN2, our global loss will be
\begin{equation}\label{eq:global-loss}
	\mathcal{L}_\text{global} = \mathbb{E}_{\mathbf{x}\sim p_\mathbf{x}} [\log D(\mathbf{x})] + \mathbb{E}_{\mathbf{z}\sim p_\mathbf{z}} [\log (1-D(G(\mathbf{z})))].
\end{equation}
where $\mathbf{x}$, $\mathbf{z}$, $p_\mathbf{x}$ and $p_{\bf z}$ are defined exactly as in the pre-trained GAN. If the generator architecture changes, it will be necessary to use the $\mathcal{L}_\text{global}$ associated to that GAN model. 

Putting everything together, the final loss function to optimize is given as
\begin{equation}\label{eq:total-loss}
	\mathcal{L} = \mathds{1}_p[\mathcal{L}_\text{local}] + \mathcal{L}_\text{global}
\end{equation}
where $\mathds{1}_p$ is an indicator function activated with probability $p$, with $p$ generally kept small to attain good convergence and stability on $G(\cdot)$. The lower the $p$, the less frequent $\mathcal{L}_\text{local}$ term updates during training. For example, when $p=1/8$, $\mathcal{L}_\text{local}$ will be updated once for every eight updates on $\mathcal{L}_\text{global}$. 

The proposed GAN inversion method is summarized in Algorithm \ref{algo:ours}. We call this algorithm {\em Clone} since the synthesized image is a near clone of the real photo.

Once this process is completed, the updated generator $G^{+}$ is used to edit the image $\mathbf{x}^*$. Editing techniques that do not require training (\eg, StyleSpace \cite{wu2021stylespace}) can be directly applied on $G^{+}$. Methods that require model training (\eg, WarpedGANSpace \cite{Tzelepis_2021_ICCV}, Image2StyleGAN \cite{Abdal2020Image2StyleGANHT}) can also be directly applied without re-training on $G^{+}$, since $G^{+}$ preserve most manifold structure as the pre-trained generator.

\begin{algorithm}[t]
    \mycaption{Pseudo-code of {\em Clone}, the proposed GAN inversion algorithm}{}
    \hspace*{\algorithmicindent} \textbf{Input:} Query image $\mathbf{x}$, pre-trained model $G(\cdot)$ and $D(\cdot)$, the training set $\mathcal{X}_{train}$, hyper-parameters $\{p,\lambda,\epsilon\}$, an inversion function $h(\cdot)$, maximum number of iterations $T$.\\
    \hspace*{\algorithmicindent} \textbf{Output:} Synthesized image $\mathbf{x}^*$, generator $G^+(\cdot)$.
    \begin{algorithmic}[1]
        \State{Let $\mathbf{z} \gets h(\mathbf{x})$.}
        \State{Let $G^{(0)}\gets G$.}
        \State{Let $D^{(0)}\gets D$.}
    	\For{$t=1$ to $T$}
            \State{$u\sim \text{Uniform}[0,1)$}
            \If{$ u < p $}
            	\State{Calculate $\mathcal{L}_\text{local}$ as in \cref{eq:local-loss}}
	    \Else
	      	\State{$\mathcal{L}_\text{local} = 0$}
	    \EndIf
            \State{Compute $\mathcal{L}_\text{global}$ as in \cref{eq:global-loss}.}
            \State{$\mathcal{L} = \mathcal{L}_\text{local} + \mathcal{L}_\text{global}$.}
            \State{$G^{(t+1)}$ is updated with $\partial \mathcal{L} / \partial G^{(t)}$.}         	
            \State{$D^{(t+1)}$ is updated with $\partial \mathcal{L} / \partial D^{(t)}$.}
            \If{$ \mathcal{L}_\text{recon} < \epsilon $} 
            	\State{Stop} 
	    \EndIf
        \EndFor{}
        \State{$G^+(\cdot) \gets G^{(t+1)}(\cdot)$.}
        \State{$\mathbf{x}^* \gets G^+({\bf z})$.}
    \end{algorithmic}\label{algo:ours}
\end{algorithm}

\section{Experiments}
\label{sec:experiments}

In this section, we provide quantitative as well as qualitative comparisons of the proposed GAN inversion algorithms against state-of-the-art techniques. 

\subsection{Implementation details}
	
We demonstrate the performance of the proposed GAN inversion algorithm on six datasets, Flickr-Faces-HQ Dataset (FFHQ)~\cite{karras2019style, karras2020analyzing}, CelebA-HQ~\cite{karras2018progressive}, Stanford Cars~\cite{Krause20133DOR}, LSUN-Cars~\cite{yu2015lsun},  Animal Faces HQ in-the-Wild (AFHQ-Wild)~\cite{choi2020stargan}, and LSUN-Horses~\cite{yu2015lsun}. The resolution of these images go from from $1,024 \times 1,024$ to $256 \times 256$ pixels.

The generator in StyleGAN2 uses a neural network architecture that receives a \emph{style} latent code at each of its layers. To achieve a disentangled latent space, the StyleGAN2 architecture employs two decoupled network components referred to as the mapping network and the synthesis network. 

A Normal distribution in $\mathcal{Z}$-space is transformed to $\mathcal{W}$-space via the mapping network, which is further extended to create $\mathcal{W+}$ \cite{Abdal2020Image2StyleGANHT}. In the experiments shown below, we recover the latent codes in $\mathcal{W+}$ space. And, in our experiments, we start with pre-trained StyleGAN2 models and apply the proposed GAN inversion algorithms detailed in Section~\ref{sec:method}. 

In our experiments, we use $p=1/4,1/8,1/32,1/16$ for the Faces (FFHQ, CelebA-HQ), Cars (LSUN-Cars, Stanford-Cars), AFHQ and LSUN-Horses experiments, respectively. We use $\lambda=10$ in \cref{eq:local-loss} for Faces, Cars and AFHQ experiments and $\lambda=20$ for LSUN-Horses experiment. These parameters are selected by a simple heuristic which is described in Supplemental File. When experimenting with additional values for these parameters, we obtained very similar results to those reported below. In all experiments, we set $\epsilon < .001$ and $T=100K$. Additional details are given in the Supplementary File. 

The starting point for ${\bf x}^*$ is given by the Restyle encoder algorithm~\cite{alaluf2021restyle}. We have also tested the use of optimization-based algorithms to initialize ${\bf x}^*$. This yielded identical results.

\begin{figure*}[h]
  \centering
  \includegraphics[width=1.0\textwidth]{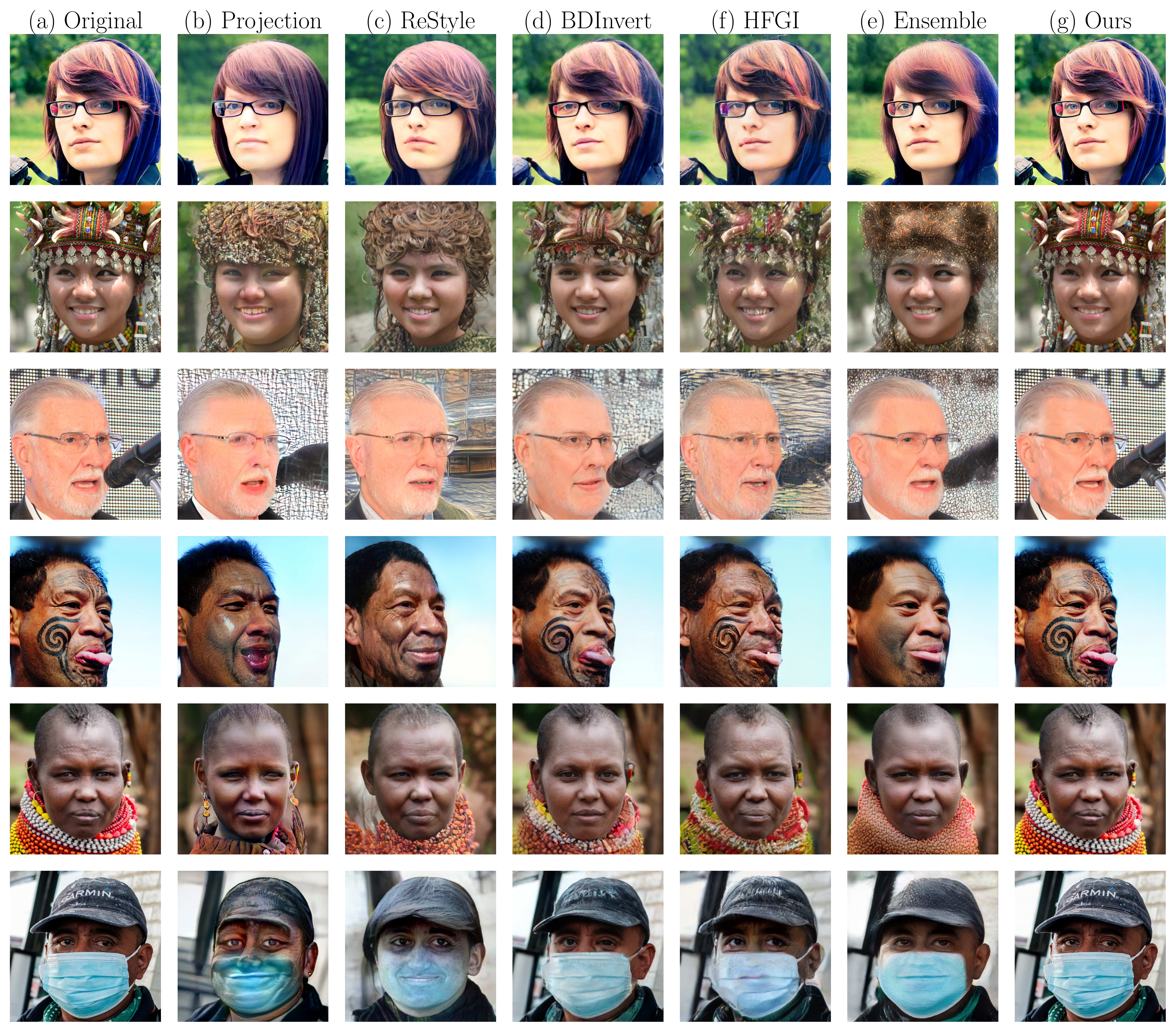}
  \caption{Comparative results of our method against five state-of-the-art GAN inversion algorithms. (a) Query photo we wish to synthesize. (b) A state-of-the-art optimization-based GAN inversion method \cite{karras2020analyzing}. (c-e) State-of-the-art encoder methods \cite{alaluf2021restyle,KANG2021GANIF,wang2021HFGI}. (f) A state-of-the-art hybrid method \cite{Chai2021EnsemblingWD}. (g) Results obtained with the approach proposed in this paper. Note how the proposed method is able to synthesize a clone that is indistinguishable from the original photo. First three rows are clones of samples found in the set used to train $G(\cdot)$. Last three rows show results on out-of-sample photos. The proposed algorithm is able to synthesize perfect clones in both, out-of-sample and in-sample, cases. All images and photos are of 1M pixels. Synthesized results best seen at 600\% zoom. }
  \vspace{-0.0em}
  \label{fig:inversion-ffhq}
\end{figure*}

\begin{figure}[h]
  \centering
  \includegraphics[width=1.0\columnwidth]{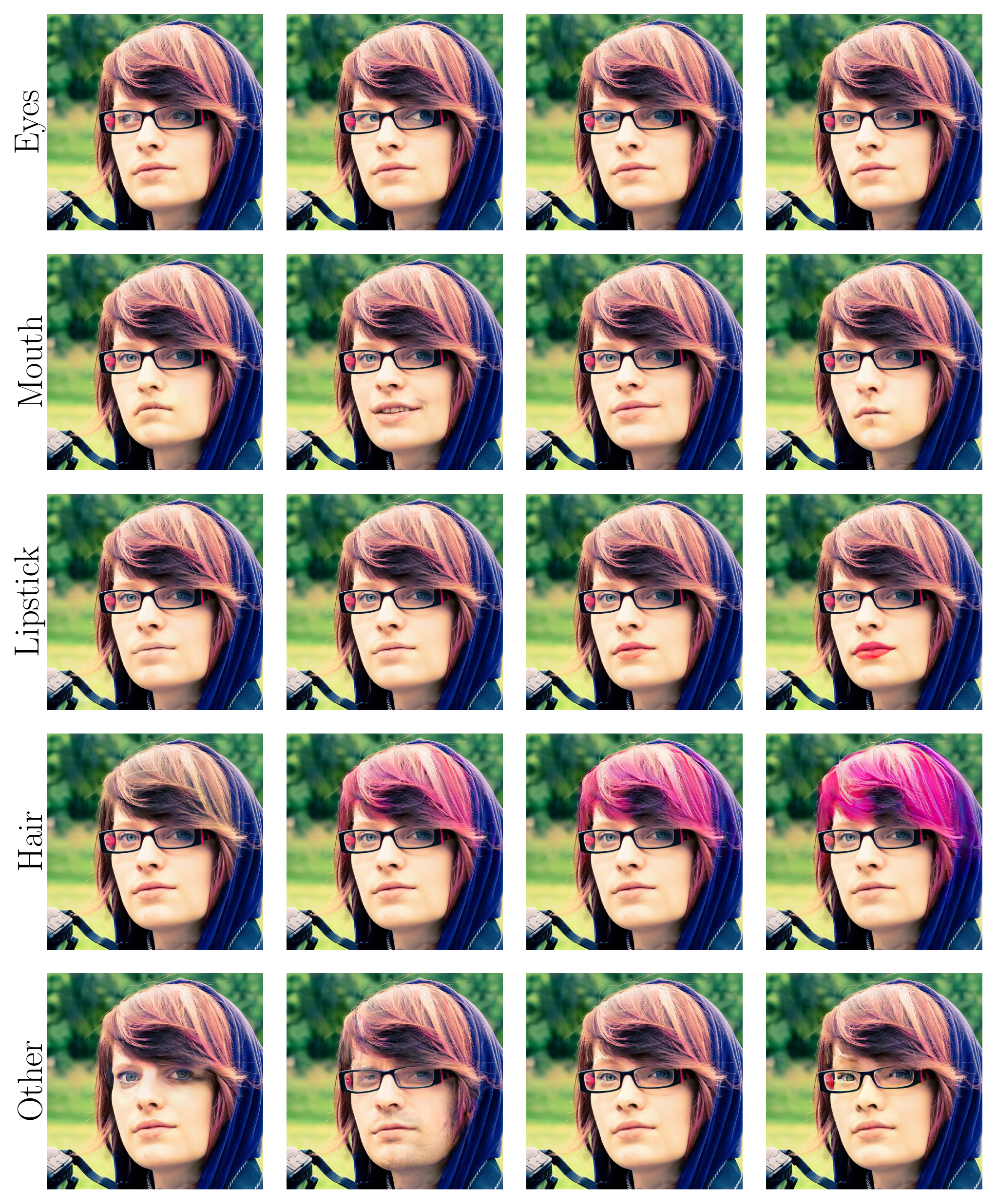}
  \caption{The synthetic clones recovered by the proposed algorithm are readily editable using standard, off-the-shelf algorithms. While other GAN inversion methods require training their own traversal functions, our proposed solution allows for the use of standard, pre-computed traversals. Here, we show eight edits using StyleSpace \cite{wu2021stylespace} on the clone of the photo shown in Figure 3(a). Note the high photo-realism of these images.}
  \label{fig:edit-ffhq}
  \vspace{-0.0em}
\end{figure}

\begin{figure*}[h]
  \centering
  \includegraphics[width=1.0\textwidth]{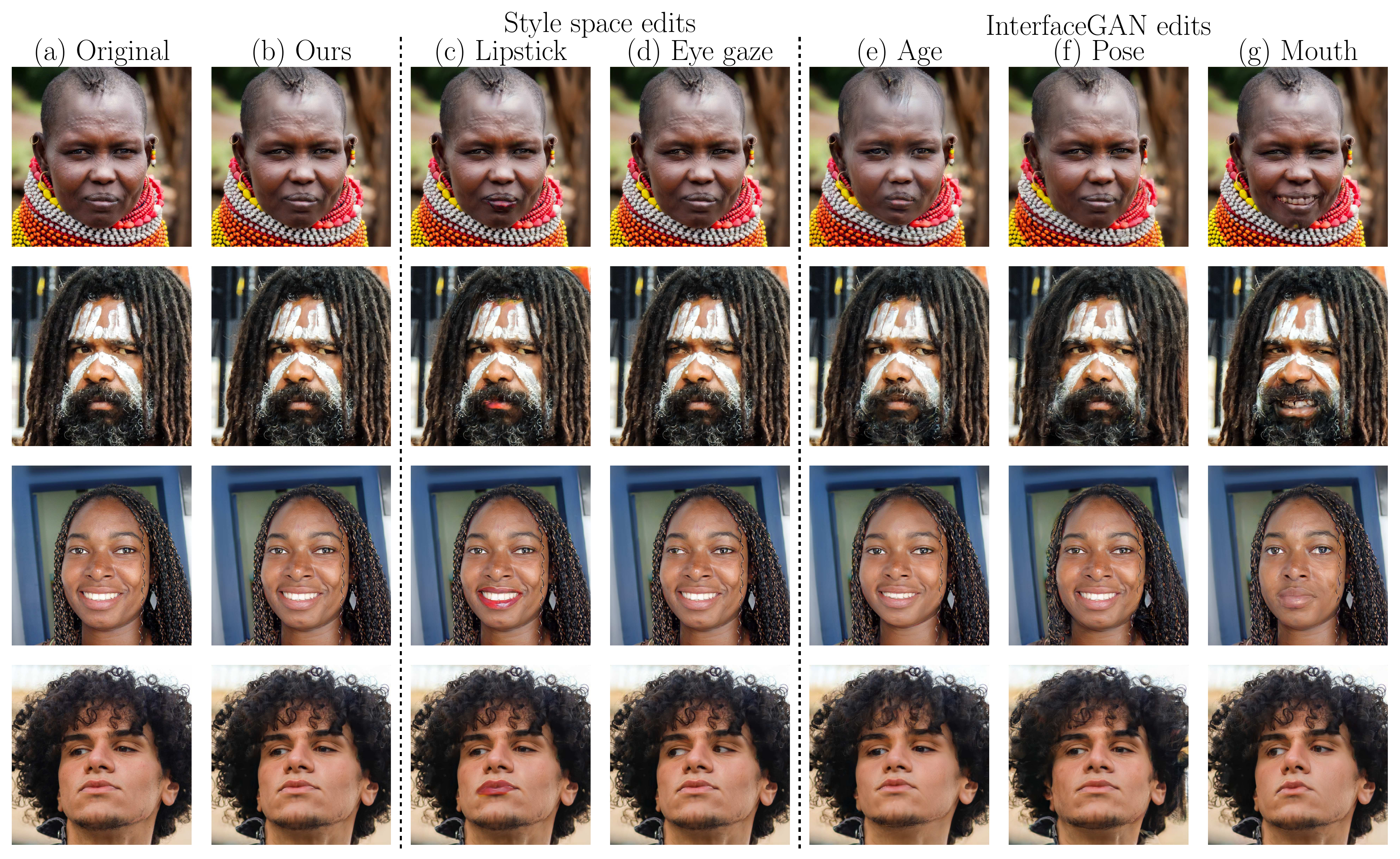}
  \caption{State-of-the-art GAN inversion algorithms cannot accurately synthesize photos of people and cultures that are under-represented in the dataset used to pre-train the GAN. This is because $G(\cdot)$ is fixed and has not been trained to represent those groups and cultures. This leads to issues of diversity and inclusion. The algorithm described in this paper addresses these limitations by allowing local tweaks to $G(\cdot)$. In (a-b), we show results of our GAN inversion algorithm on photos of people with hair styles as well as facial scarification, tattoos, and cultural marks not well represented or not represented at all in the set used to pre-train the GAN. In (c-d), we show edits of the images synthesized by our approach using the algorithm of \cite{wu2021stylespace}. (e-g) shows edits using the algorithm of \cite{shen2020interfacegan}.}
  \vspace{-1em}
  \label{fig:inclusive}
\end{figure*}

\subsection{Qualitative results on human faces} 

We first show our algorithm is able to solve the inversion problem of the target photos with extreme accuracy, even for out-of-sample photos, \cref{fig:inversion-ffhq}. The figure provides  comparative results against Restyle~\cite{alaluf2021restyle}, BDInvert~\cite{KANG2021GANIF}, High-Fidelity GAN Inversion (HFGI) \cite{wang2021HFGI}, and Ensemble (a hybrid optimization+encoder approach) \cite{Chai2021EnsemblingWD}. 

As seen in \cref{fig:inversion-ffhq}, the proposed algorithm is able to synthesize image clones that are basically identical to the real photos. The synthesized images include difficult-to-reconstruct, high-frequency details such as hair and skin texture. (Figure results best seen at a 600\% zoom.)

Second, to demonstrate that such accurate reconstructions come at no adverse effect on the ability to edit the synthetic clone, we show a number of image modifications using standard algorithms~\cite{wu2021stylespace,shen2020interfacegan}, \cref{fig:edit-ffhq}. 

Specifically, we chose two state-of-the-art GAN editing techniques. The first is StyleSpace~\cite{wu2021stylespace}, which discovers dimensions in latent space that control specific image attributes such as eye gaze, hair color, skin tone, and mouth shape. The second is InterfaceGAN~\cite{shen2020interfacegan}, which finds linear traversals that modify a specific semantic attribute.

\subsection{Diversity and inclusion}

The biases of computer vision and machine learning models are well known and widely reported \cite{raji2019actionable,torralba2011unbiased,beery2018recognition}. These are especially problematic when dealing with human faces due to our attachment of self-worth, identity and cultural values \cite{menon2020pulse}. GANs are not immune from this problem, with pre-trained models carrying the biases of the photos used to define their training sets \cite{jo2020lessons,balakrishnan2020towards}. 

The GAN inversion algorithm derived in this paper solves this important problem. Our algorithm is especially good at synthesizing  out-of-sample data points. This means, the proposed algorithm successfully synthesizes photos of people and cultures under-represented in the training set. In fact, the proposed inversion algorithm is still able to synthesize a near identical images to their query photos. 

\cref{fig:inclusive} shows several example image clones corresponding to out-of-sample photos. Note how the proposed algorithm is able to synthesize hairstyle, facial tattoos, and cultural jewelry/amulets not included in the training set of the pre-trained GAN model we used. 

Importantly, and as shown in the figure, these clones are equally editable to those of in-sample groups and cultures.

\subsection{Qualitative results on cars and animals} 

\begin{figure*}[h]
  \centering
  \includegraphics[width=1.0\textwidth]{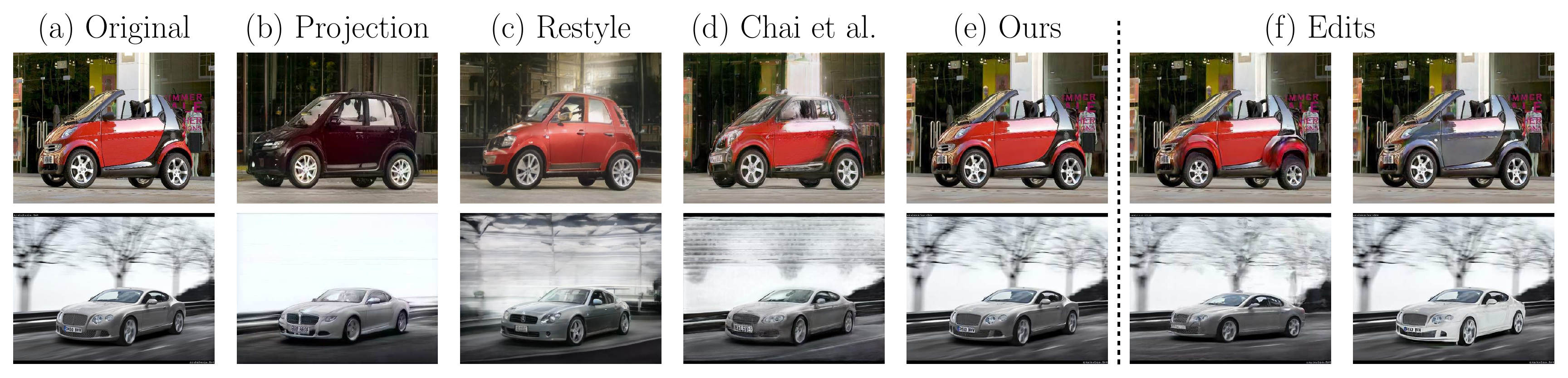}
  \caption{Inversion results on Stanford Cars \cite{Krause20133DOR}. (a) Original photo. (b-d) State-of-the-art results. (e) Our results. (f) A couple edits on our reconstruction using \cite{wu2021stylespace} and \cite{shen2021closedform}.}
  \label{fig:cars}
\end{figure*}

\cref{fig:cars} shows qualitative results for cars on the Stanford Cars \cite{Krause20133DOR} dataset using StyleGAN2 models pre-trained on LSUN-Cars dataset \cite{yu2015lsun}. In (a) we show the query photo we wish to synthesized. Results obtained with state-of-the-art methods are in (b-d). The synthesis results given by our algorithm are in (e). And, in (f), we show a couple edits applied to our synthesized images. As with faces, we see that the proposed approach yields results that are indistinguishable from the original photo.

\begin{figure*}[h]
  \centering
  \includegraphics[width=1.0\textwidth]{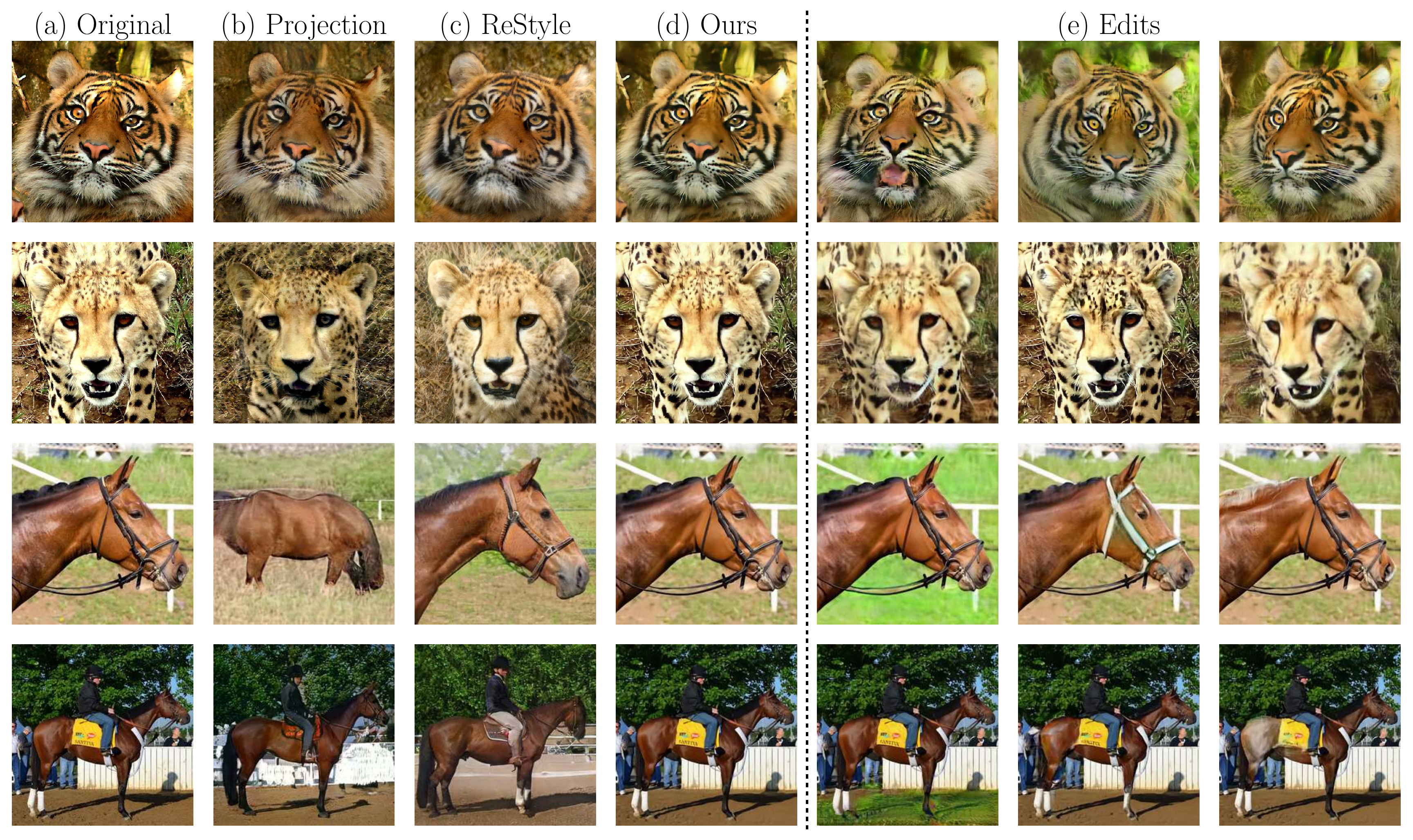}
  \caption{Reconstruction results of the real photos shown in column (a). Projection (an optimization method) \cite{karras2020analyzing} results are in column (b), ReStyle (an Encoder-based method) \cite{alaluf2021restyle} results in column (c), and the results of the approach described in this paper in column (d). In (e), we show three image edits using StyleSpace \cite{wu2021stylespace} and the unsupervised algorithm of \cite{shen2021closedform}. The first two rows show results on the AFHQ-Wild database \cite{choi2020stargan}. The last two rows show results on the LSUN Horse dataset \cite{yu2015lsun}.}
  \label{fig:afhq-lsunhorse}
\end{figure*}

\cref{fig:afhq-lsunhorse} shows GAN inversion results of animals on the AFHQ-Wild and LSUN horses datasets. As with faces and cars, all StyleGAN2 models were pre-trained on the same training set and the results shown in the figure computed on an independent set of photos. These two datasets only have pre-train models for two of the GAN inversion models, Projection \cite{karras2020analyzing} and ReStyle \cite{alaluf2021restyle}. Thus, \cref{fig:afhq-lsunhorse}(a) shows the original photo, \cref{fig:afhq-lsunhorse} (b-c) these state-of-the-art results, and \cref{fig:afhq-lsunhorse}(d) the reconstructions given by the {\em Clone} algorithm derived in this paper. In (e), we show three example edits using StyleSpace \cite{wu2021stylespace} and the unsupervised approach of \cite{shen2021closedform}.

\subsection{Quantitative results} 

Previous sections provided a number of qualitative comparative results on faces, cars and animals. 

Table \ref{tab:fid_scores} shows the corresponding quantitative results. The first results in this table are computed on the CelebA-HQ~\cite{karras2018progressive} face dataset using StyleGAN2 models pre-trained on FFHQ \cite{karras2019style}. The second results are on the testing set of the AFHQ-Wild \cite{choi2020stargan} animal dataset using StyleGAN2 models pre-trained on the training set of the same database. The third results are on the testing set of the Stanford Cars \cite{Krause20133DOR} dataset using StyleGAN2 models pre-trained on the the training set of the same database. 

We used two metrics to report our quantitative results in Table \ref{tab:fid_scores}. The first is the Mean Squared Error (MSE). This is simply the norm-2 distance between the query photo and the image synthesized by each of the five state-of-the-art methods plus the algorithm presented in this paper. Results are the average over all testing images. Obviously, the lower the MSE, the better, with zero indicating the original photos and the synthesized clones are 100\% identical. The second metric is the Learned Perceptual Image Patch Similarity (LPIPS) \cite{zhang2018unreasonable}. LPIPS computes the perceptual similarity of the synthesized image to the query photo. The perceptual similarity is calculated using a visual neural network model. As it is most commonly done, we use VGG \cite{Simonyan15}. As with MSE, the lower the value of LPIPS, the more visually similar the synthesized images are to the query photos.

As we see in the table, our proposed algorithm achieves MSE values that are at least an order of magnitude lower than those obtained by state-of-the-art methods. This is true regardless of the database and type of object we wish to synthesize. LPIPS confirms the visual similarity of our synthesized images to their corresponding query photos. Not all GAN inversion algorithms have a model available for comparison. When that's the case, we indicate this with `--' entries in the table.

We refer the reader to the Supplementary Files for additional quantitative and qualitative results.

\begin{table*}[!t]
	\small{
		\begin{center}
		\newcommand{\tabwid}{0.10cm}
		\newcommand{\expwid}{1.45cm}
		\renewcommand{\arraystretch}{1}
		\begin{threeparttable}
			\begin{tabular}{p{2.15cm}>{\centering\arraybackslash}p{\expwid}>{\centering\arraybackslash}p{\expwid}>{\centering\arraybackslash}p{\expwid}>{\centering\arraybackslash}p{\expwid}>{\centering\arraybackslash}p{\expwid}>{\centering\arraybackslash}p{\expwid}>{\centering\arraybackslash}p{\expwid}>{\centering\arraybackslash}p{\expwid}>{\centering\arraybackslash}p{\expwid}}
				\toprule
				Dataset & \multicolumn{2}{c}{\small{CelebA-HQ}}  & \multicolumn{2}{c}{\small{Stanford Cars}} &  \multicolumn{2}{c}{\small{AFHQ-Wild}} &  \multicolumn{2}{c}{\small{LSUN-Horse}}\\ 
				\cmidrule{1-1} \cmidrule(lr){2-3} \cmidrule(lr){4-5} \cmidrule(lr){6-7} \cmidrule(lr){8-9}
				Metric 									& MSE 				& LPIPS 					& MSE				& LPIPS				& MSE 				& LPIPS  			& MSE 			& LPIPS \\ 
				\cmidrule{1-1} \cmidrule(lr){2-3} \cmidrule(lr){4-5} \cmidrule(lr){6-7} \cmidrule(lr){8-9}
				Projection	$^\text{O}$ \cite{karras2020analyzing}	& .074 $\pm$.055 		& .429 $\pm$.044   			& .318 $\pm$.120  		& .486 $\pm$.067		& .126 $\pm$.066		& .491 $\pm$.036		& .240 $\pm$.195  	& .454 $\pm$.072	\\ 
				\cmidrule{1-1}
				ReStyle $^\text{E}$	\cite{alaluf2021restyle}		& .050 $\pm$.019 		& .475 $\pm$.038   			& .082 $\pm$.035  		& .352 $\pm$.063	& .085 $\pm$.039 		& .509 $\pm$.037 		& .159 $\pm$.070  		& .525 $\pm$.071 	\\ 
				\cmidrule{1-1}
				BDInvert $^\text{E}$	\cite{KANG2021GANIF}		& .016 $\pm$.080 		& .373 $\pm$.040   			& --					& -- 					& --  					& -- 			& --  					& --	\\ 
				\cmidrule{1-1}
				HFGI $^\text{E}$ \cite{wang2021HFGI}			& .032 $\pm$.054		& .423 $\pm$.045      		& --					& --					& --  					& -- 			& --  					& --	\\ 
				\cmidrule{1-1}
				Ensemble $^\text{H}$ \cite{Chai2021EnsemblingWD}& .017 $\pm$.011 		& .373 $\pm$.038   			& .284 $\pm$.025  		& .448 $\pm$.053		& --  					& -- 					& --  					& --\\ 
				\hline
				Ours 									& {\bf .004} $\pm$.006 	& {\bf .283} $\pm$.050 		& {\bf .006} $\pm$.007    	& {\bf .154} $\pm$.046 	& {\bf .014} $\pm$.013 	& {\bf .382} $\pm$.087 	& {\bf .005} $\pm$.009    	& {\bf .141} $\pm$.043\\ 
				\bottomrule
			\end{tabular}
	\end{threeparttable}
	\end{center}}
	  \vspace{-1.2em}
	\caption{Quantitative metrics of the proposed algorithm compared to results given by state-of-the-art baseline methods. MSE is the Mean Squared Error between the query photo and its synthesized copy as given by each of the listed algorithms. LPIPS (Learned Perceptual Image Patch Similarity) \cite{zhang2018unreasonable} computes the perceptual similarity between the original photo and the synthesized clones using a VGG network. LPIPS and MSE are averaged across samples. The GAN inversion method (optimization, encoder, hybrid) is specified as: $^\text{E}$: encoder-based, $^\text{O}$: optimization-based, $^\text{H}$: hybrid method. `--' indicates that the pre-trained models for the corresponding datasets are not publicly available.}
	\label{tab:fid_scores}
\end{table*}

\section{Assumptions and Limitation}\label{sec:limitation}

Additionally, the almost perfect GAN inversion results shown above come at an additional small computational cost compared to previous methods. Since previous algorithms optimize $h(\cdot)$, their cost is associated to the number of iterations required to get good convergence. The approach proposed in this paper locally tweaks $G(\cdot)$. Note that this is {\em not} the same as re-training the GAN model. This local tweak is successfully completed in just a few iterations, taking typically several seconds to a few minutes. In the worse cases, where our algorithm's solution ${\bf x}^*$ is significantly far from $\hat{\bf x}$, the {\em Clone} algorithm may take several minutes. This may limit the use of the proposed technique in applications that require close to real-time results.

\section{Conclusion}
\label{sec:conclusion}

GAN inversion is a hard problem, with limitations on which photos can and cannot be synthesized based on GAN architectures, loss functions, and training sets, among others. Here, we have defined a solution to these limitations by allowing the generator function $G(\cdot)$ to be locally updated to include the image we wish to synthesize. Using extensive experimental results, we have shown that the proposed approach yields near perfect real photo reconstruction that are editable, with quantitative evaluations yielding results an order of magnitude (or more) better than current state-of-the-art algorithms. 


{\small
\bibliographystyle{ieee_fullname}
\bibliography{references}

\begin{thebibliography}{10}\itemsep=-1pt

\bibitem{Abdal2019Image2StyleGANHT}
Rameen Abdal, Yipeng Qin, and Peter Wonka.
\newblock Image2stylegan: How to embed images into the stylegan latent space?
\newblock {\em 2019 IEEE/CVF International Conference on Computer Vision
  (ICCV)}, pages 4431--4440, 2019.

\bibitem{Abdal2020Image2StyleGANHT}
Rameen Abdal, Yipeng Qin, and Peter Wonka.
\newblock Image2stylegan++: How to edit the embedded images?
\newblock {\em CVPR}, pages 8293--8302, 2020.

\bibitem{adelson1984pyramid}
Edward~H Adelson, Charles~H Anderson, James~R Bergen, Peter~J Burt, and Joan~M
  Ogden.
\newblock Pyramid methods in image processing.
\newblock {\em RCA engineer}, 29(6):33--41, 1984.

\bibitem{alaluf2021restyle}
Yuval Alaluf, Or Patashnik, and Daniel Cohen-Or.
\newblock Restyle: A residual-based stylegan encoder via iterative refinement.
\newblock In {\em Proceedings of the IEEE/CVF International Conference on
  Computer Vision (ICCV)}, October 2021.

\bibitem{balakrishnan2020towards}
Guha Balakrishnan, Yuanjun Xiong, Wei Xia, and Pietro Perona.
\newblock Towards causal benchmarking of bias in face analysis algorithms.
\newblock {\em arXiv e-prints}, pages arXiv--2007, 2020.

\bibitem{Bau2019SeeingWA}
David Bau, Jun-Yan Zhu, Jonas Wulff, William~S. Peebles, Hendrik Strobelt,
  Bolei Zhou, and Antonio Torralba.
\newblock Seeing what a gan cannot generate.
\newblock {\em ICCV}, pages 4501--4510, 2019.

\bibitem{beery2018recognition}
Sara Beery, Grant Van~Horn, and Pietro Perona.
\newblock Recognition in terra incognita.
\newblock In {\em Proceedings of the European conference on computer vision
  (ECCV)}, pages 456--473, 2018.

\bibitem{bojanowski2018optimizing}
Piotr Bojanowski, Armand Joulin, David Lopez-Pas, and Arthur Szlam.
\newblock Optimizing the latent space of generative networks.
\newblock In {\em International Conference on Machine Learning}, pages
  600--609. PMLR, 2018.

\bibitem{brock2018large}
Andrew Brock, Jeff Donahue, and Karen Simonyan.
\newblock Large scale gan training for high fidelity natural image synthesis.
\newblock In {\em International Conference on Learning Representations}, 2018.

\bibitem{Chai2021UsingLS}
Lucy Chai, Jonas Wulff, and Phillip Isola.
\newblock Using latent space regression to analyze and leverage
  compositionality in gans.
\newblock {\em ICLR}, 2021.

\bibitem{Chai2021EnsemblingWD}
Lucy Chai, Jun-Yan Zhu, Eli Shechtman, Phillip Isola, and Richard Zhang.
\newblock Ensembling with deep generative views.
\newblock In {\em CVPR}, 2021.

\bibitem{choi2020stargan}
Yunjey Choi, Youngjung Uh, Jaejun Yoo, and Jung-Woo Ha.
\newblock Stargan v2: Diverse image synthesis for multiple domains.
\newblock In {\em Proceedings of the IEEE/CVF Conference on Computer Vision and
  Pattern Recognition}, pages 8188--8197, 2020.

\bibitem{Creswell2019InvertingTG}
Antonia Creswell and Anil~Anthony Bharath.
\newblock Inverting the generator of a generative adversarial network.
\newblock {\em IEEE Transactions on Neural Networks and Learning Systems},
  30:1967--1974, 2019.

\bibitem{Gadde_2021_ICCV}
Raghudeep Gadde, Qianli Feng, and Aleix~M. Martinez.
\newblock Detail me more: Improving gan's photo-realism of complex scenes.
\newblock In {\em Proceedings of the IEEE/CVF International Conference on
  Computer Vision (ICCV)}, pages 13950--13959, October 2021.

\bibitem{jo2020lessons}
Eun~Seo Jo and Timnit Gebru.
\newblock Lessons from archives: Strategies for collecting sociocultural data
  in machine learning.
\newblock In {\em Proceedings of the 2020 Conference on Fairness,
  Accountability, and Transparency}, pages 306--316, 2020.

\bibitem{KANG2021GANIF}
Kyoungkook Kang, Seongtae Kim, and Sunghyun Cho.
\newblock Gan inversion for out-of-range images with geometric transformations.
\newblock {\em ICCV}, 2021.

\bibitem{karras2018progressive}
Tero Karras, Timo Aila, Samuli Laine, and Jaakko Lehtinen.
\newblock Progressive growing of gans for improved quality, stability, and
  variation.
\newblock In {\em International Conference on Learning Representations}, 2018.

\bibitem{Karras2021}
Tero Karras, Miika Aittala, Samuli Laine, Erik H\"ark\"onen, Janne Hellsten,
  Jaakko Lehtinen, and Timo Aila.
\newblock Alias-free generative adversarial networks.
\newblock In {\em Proc. NeurIPS}, 2021.

\bibitem{karras2019stylebased}
Tero Karras, Samuli Laine, and Timo Aila.
\newblock A style-based generator architecture for generative adversarial
  networks, 2019.

\bibitem{karras2019style}
Tero Karras, Samuli Laine, and Timo Aila.
\newblock A style-based generator architecture for generative adversarial
  networks.
\newblock In {\em Proceedings of the IEEE conference on computer vision and
  pattern recognition}, pages 4401--4410, 2019.

\bibitem{Karras2019stylegan2}
Tero Karras, Samuli Laine, Miika Aittala, Janne Hellsten, Jaakko Lehtinen, and
  Timo Aila.
\newblock Analyzing and improving the image quality of {StyleGAN}.
\newblock In {\em Proc. CVPR}, 2020.

\bibitem{karras2020analyzing}
Tero Karras, Samuli Laine, Miika Aittala, Janne Hellsten, Jaakko Lehtinen, and
  Timo Aila.
\newblock Analyzing and improving the image quality of stylegan.
\newblock In {\em Proceedings of the IEEE/CVF Conference on Computer Vision and
  Pattern Recognition}, pages 8110--8119, 2020.

\bibitem{Krause20133DOR}
Jonathan Krause, Michael Stark, Jia Deng, and Li Fei-Fei.
\newblock 3d object representations for fine-grained categorization.
\newblock {\em 2013 IEEE International Conference on Computer Vision
  Workshops}, pages 554--561, 2013.

\bibitem{Lipton2017PreciseRO}
Zachary~Chase Lipton and Subarna Tripathi.
\newblock Precise recovery of latent vectors from generative adversarial
  networks.
\newblock {\em ArXiv}, abs/1702.04782, 2017.

\bibitem{menon2020pulse}
Sachit Menon, Alexandru Damian, Shijia Hu, Nikhil Ravi, and Cynthia Rudin.
\newblock Pulse: Self-supervised photo upsampling via latent space exploration
  of generative models.
\newblock In {\em Proceedings of the ieee/cvf conference on computer vision and
  pattern recognition}, pages 2437--2445, 2020.

\bibitem{pidhorskyi2020adversarial}
Stanislav Pidhorskyi, Donald~A Adjeroh, and Gianfranco Doretto.
\newblock Adversarial latent autoencoders.
\newblock In {\em Proceedings of the IEEE/CVF Conference on Computer Vision and
  Pattern Recognition}, pages 14104--14113, 2020.

\bibitem{raji2019actionable}
Inioluwa~Deborah Raji and Joy Buolamwini.
\newblock Actionable auditing: Investigating the impact of publicly naming
  biased performance results of commercial ai products.
\newblock In {\em Proceedings of the 2019 AAAI/ACM Conference on AI, Ethics,
  and Society}, pages 429--435, 2019.

\bibitem{Richardson2020EncodingIS}
Elad Richardson, Yuval Alaluf, Or Patashnik, Yotam Nitzan, Yaniv Azar, Stav
  Shapiro, and Daniel Cohen-Or.
\newblock Encoding in style: a stylegan encoder for image-to-image translation.
\newblock {\em CVPR}, 2021.

\bibitem{shen2020interfacegan}
Yujun Shen, Ceyuan Yang, Xiaoou Tang, and Bolei Zhou.
\newblock Interfacegan: Interpreting the disentangled face representation
  learned by gans.
\newblock {\em IEEE transactions on pattern analysis and machine intelligence},
  2020.

\bibitem{shen2021closedform}
Yujun Shen and Bolei Zhou.
\newblock Closed-form factorization of latent semantics in gans.
\newblock In {\em CVPR}, 2021.

\bibitem{Simonyan15}
Karen Simonyan and Andrew Zisserman.
\newblock Very deep convolutional networks for large-scale image recognition.
\newblock In {\em International Conference on Learning Representations}, 2015.

\bibitem{torralba2011unbiased}
Antonio Torralba and Alexei~A Efros.
\newblock Unbiased look at dataset bias.
\newblock In {\em CVPR 2011}, pages 1521--1528. IEEE, 2011.

\bibitem{Tzelepis_2021_ICCV}
Christos Tzelepis, Georgios Tzimiropoulos, and Ioannis Patras.
\newblock {WarpedGANSpace}: Finding non-linear rbf paths in {GAN} latent space.
\newblock In {\em Proceedings of the IEEE/CVF International Conference on
  Computer Vision (ICCV)}, pages 6393--6402, October 2021.

\bibitem{wang2021HFGI}
Tengfei Wang, Yong Zhang, Yanbo Fan, Jue Wang, and Qifeng Chen.
\newblock High-fidelity gan inversion for image attribute editing.
\newblock {\em arxiv:2109.06590}, 2021.

\bibitem{wu2021stylespace}
Zongze Wu, Dani Lischinski, and Eli Shechtman.
\newblock Stylespace analysis: Disentangled controls for stylegan image
  generation.
\newblock In {\em Proceedings of the IEEE/CVF Conference on Computer Vision and
  Pattern Recognition}, pages 12863--12872, 2021.

\bibitem{xia2021gan}
Weihao Xia, Yulun Zhang, Yujiu Yang, Jing-Hao Xue, Bolei Zhou, and Ming-Hsuan
  Yang.
\newblock Gan inversion: A survey.
\newblock {\em arXiv preprint arXiv:2101.05278}, 2021.

\bibitem{yu2015lsun}
Fisher Yu, Ari Seff, Yinda Zhang, Shuran Song, Thomas Funkhouser, and Jianxiong
  Xiao.
\newblock Lsun: Construction of a large-scale image dataset using deep learning
  with humans in the loop.
\newblock {\em arXiv preprint arXiv:1506.03365}, 2015.

\bibitem{zhang2018unreasonable}
Richard Zhang, Phillip Isola, Alexei~A Efros, Eli Shechtman, and Oliver Wang.
\newblock The unreasonable effectiveness of deep features as a perceptual
  metric.
\newblock In {\em Proceedings of the IEEE conference on computer vision and
  pattern recognition}, pages 586--595, 2018.

\bibitem{Zhu2016GenerativeVM}
Jun-Yan Zhu, Philipp Kr{\"a}henb{\"u}hl, Eli Shechtman, and Alexei~A. Efros.
\newblock Generative visual manipulation on the natural image manifold.
\newblock In {\em ECCV}, 2016.

\end{thebibliography}


\begin{thebibliography}{1}\itemsep=-1pt

\bibitem{wu2021stylespace}
Zongze Wu, Dani Lischinski, and Eli Shechtman.
\newblock Stylespace analysis: Disentangled controls for stylegan image
  generation.
\newblock In {\em Proceedings of the IEEE/CVF Conference on Computer Vision and
  Pattern Recognition}, pages 12863--12872, 2021.

\end{thebibliography}
}

\end{document}


\title{Supplemental Material for Near Perfect GAN Inversion}
\twocolumn[{\renewcommand\twocolumn[1][]{#1}%
\maketitle

\vspace{-1.1cm}
\begin{center}
    \includegraphics[width=0.95\textwidth]{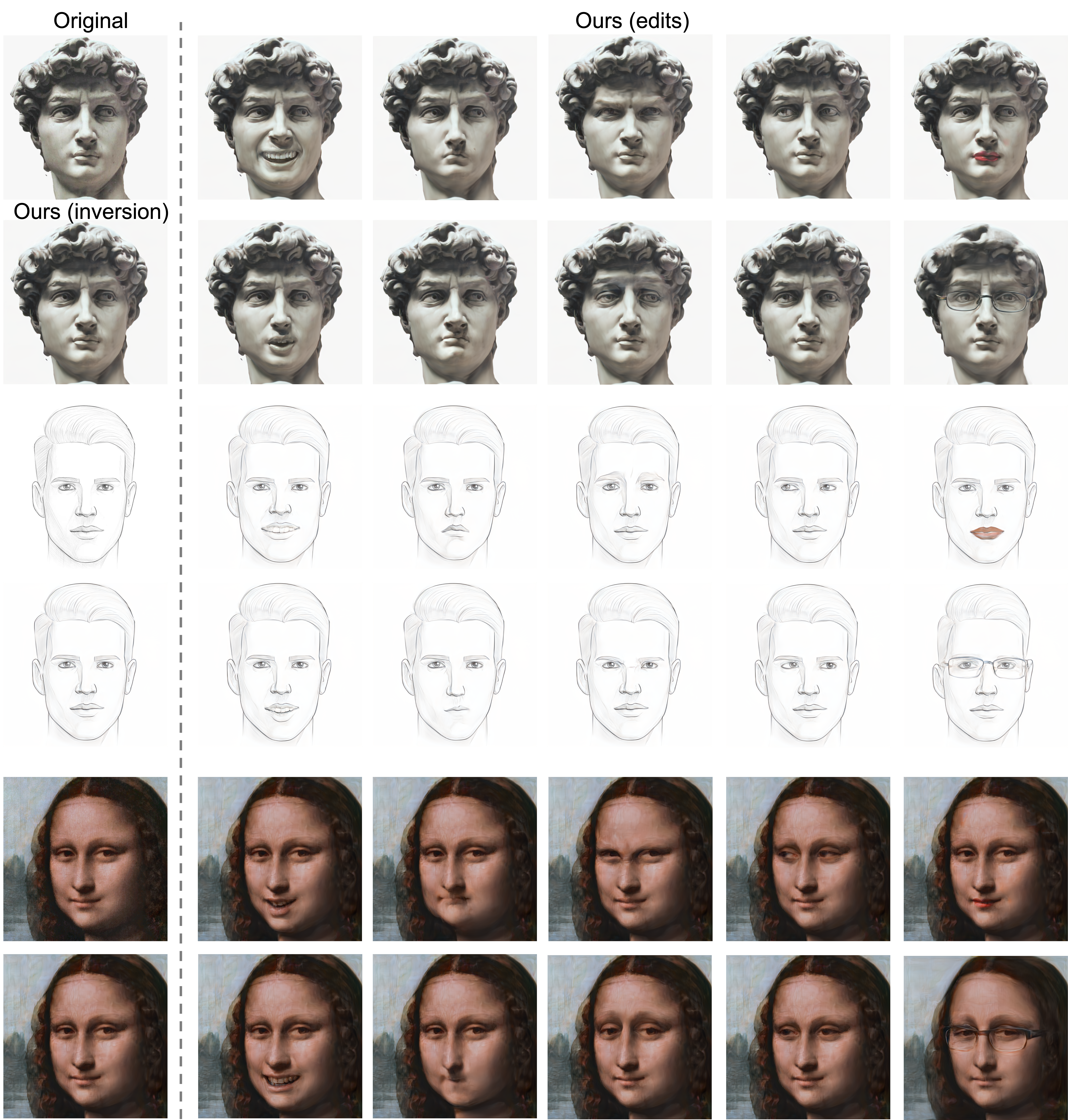}
\end{center}
\vspace{-.5cm}
\captionof{figure}{Clones and edits using a SytleGAN2 model pre-trained on FFHQ.  Left most column: Shown here are the originals and synthetic clones given by our proposed algorithm of a picture of Michelangelo’s statue of David, a drawing (sketch) of a human face, and Da Vinci’s Mona Lisa. Other columns: Edits of our synthetic clones using the methods of \cite{wu2021stylespace}. }
\label{fig:1}
\vspace{.25cm}
}]

\begin{figure*}[h!]
  \centering
  \includegraphics[width=0.85\textwidth]{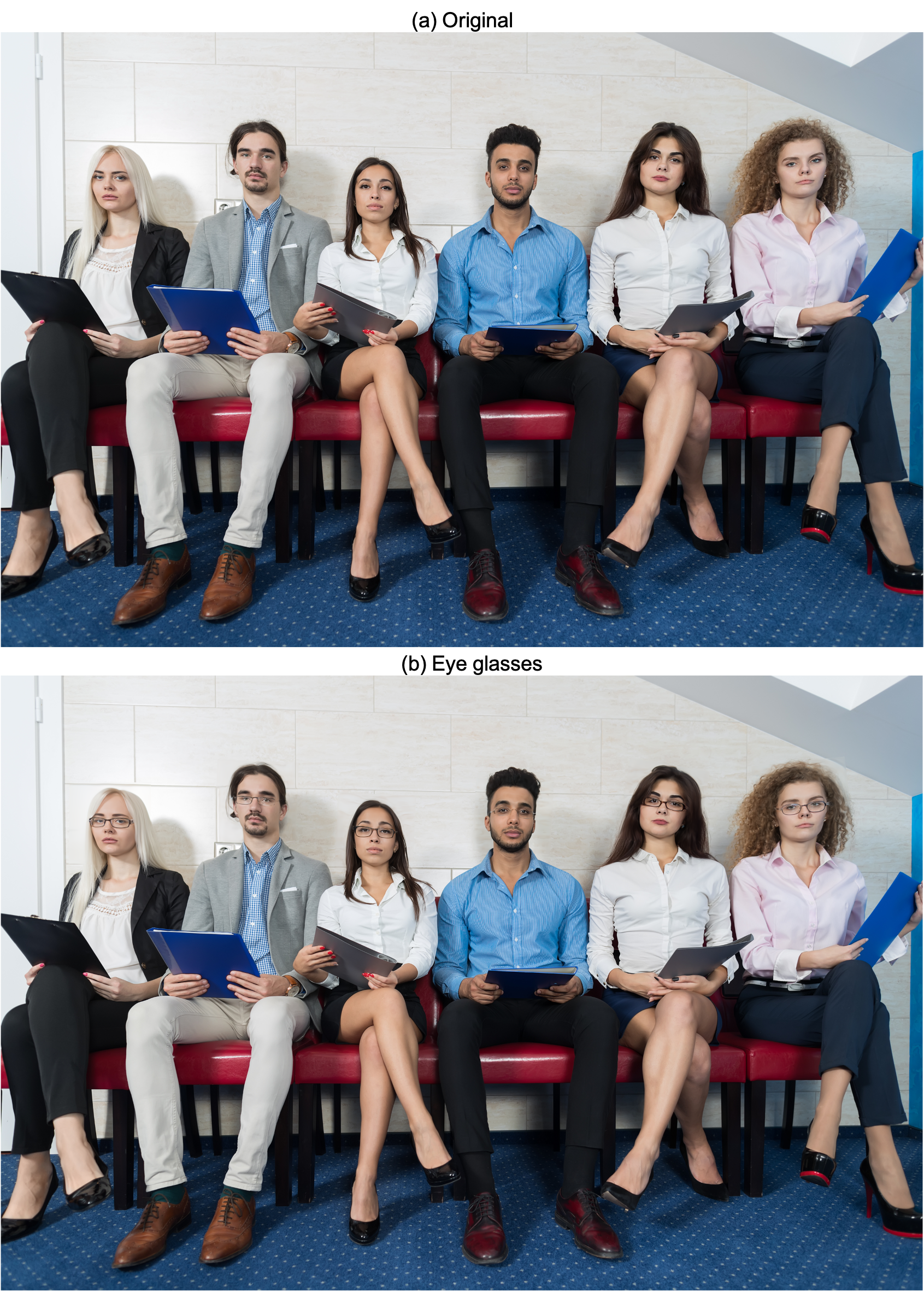}
  \caption{The proposed algorithm can be applied to images with multiple faces. Here, we use a face detector and a facial landmark detector to crop and align all faces. Next, our Clone algorithm is used to synthesize these faces, followed by edits given by off-the-shelf algorithms. Finally, the edited synthetic images are added to the original photo. (a) Original photo. (b) In this image, we have substituted the original faces by their synthetic clones edited to include eye glasses.}
  \label{fig:2}
\end{figure*}
\renewcommand{\thefigure}{S\arabic{figure} (Cont.)}
\addtocounter{figure}{-1}
\begin{figure*}[h!]
  \centering
  \includegraphics[width=0.85\textwidth]{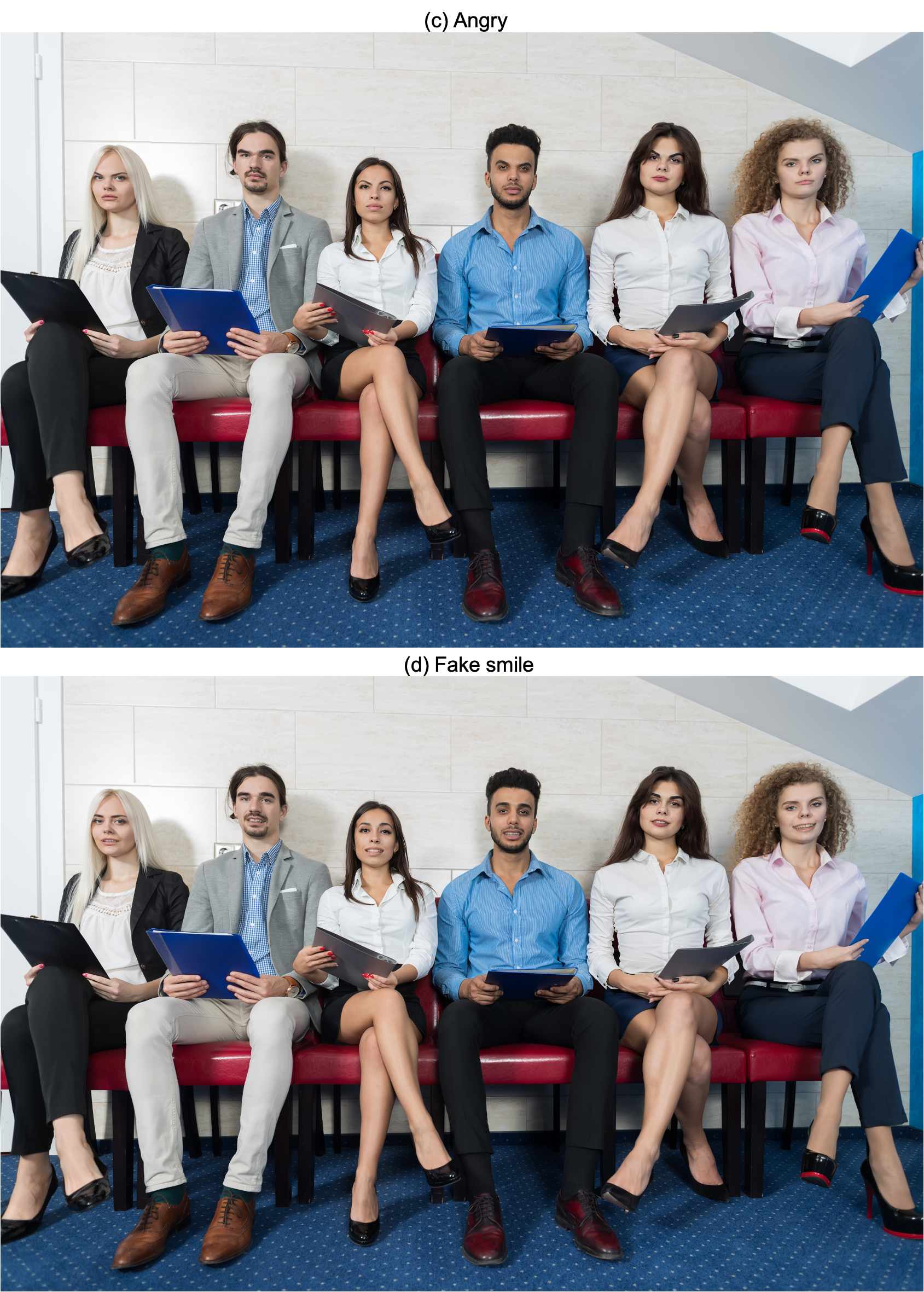}
  \caption{(c) Here, we have edited all faces to look angry. (d) And, in this example, we have edited all faces to have an open mouth (showing teeth), giving the impression of a ‘fake’ smile.}
  \label{fig:3}
\end{figure*}
\addtocounter{figure}{-1}
\begin{figure*}[t!]
  \centering
  \includegraphics[width=0.85\textwidth]{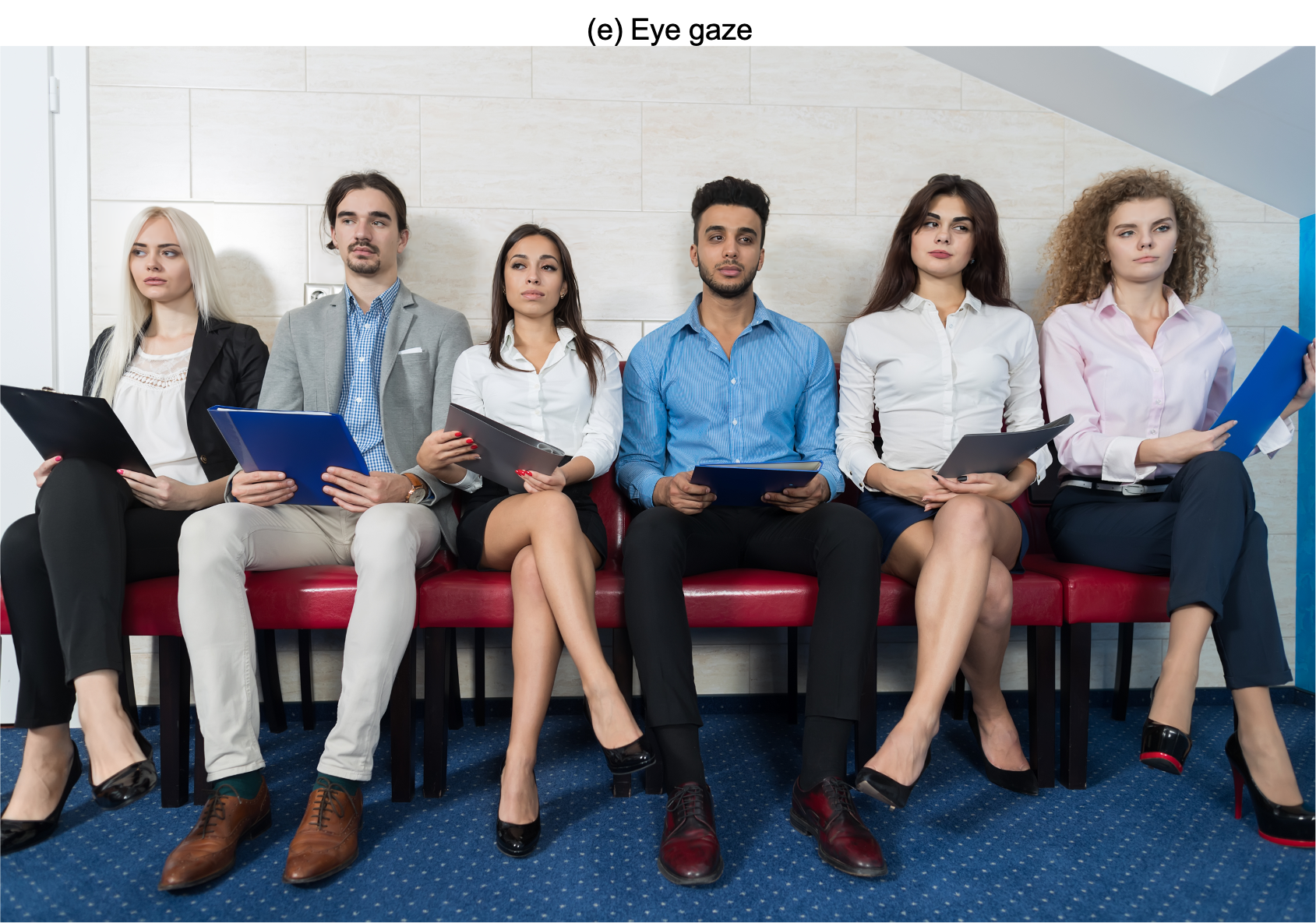}
  \caption{(e) In this last example, we have edited the eye gaze direction of the individuals in the photo, giving an impression of being distracted.}
  \label{fig:4}
\end{figure*}

\renewcommand{\thefigure}{S\arabic{figure}}
\newpage
\vspace{-4cm}
\begin{figure*}[h]
  \centering
  \includegraphics[width=0.85\textwidth]{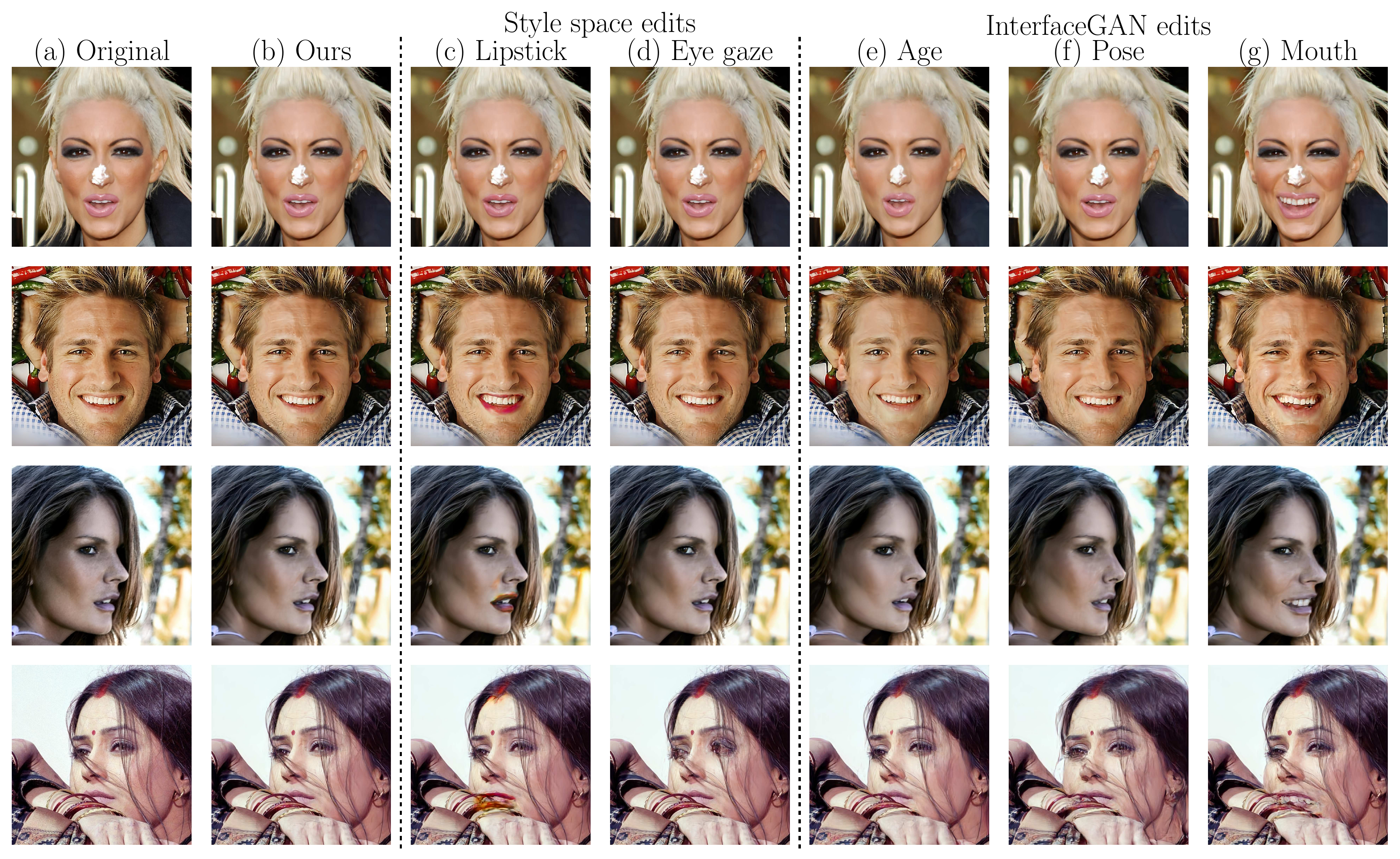}
  \caption{Additional qualitative results of our proposed Clone algorithm on images of Celeb-HQ using a StyleGAN2 model trained on FFHQ.}
  \label{fig:5}
\end{figure*}

\newpage
\begin{figure*}[h]
  \centering
  \includegraphics[width=1.0\textwidth]{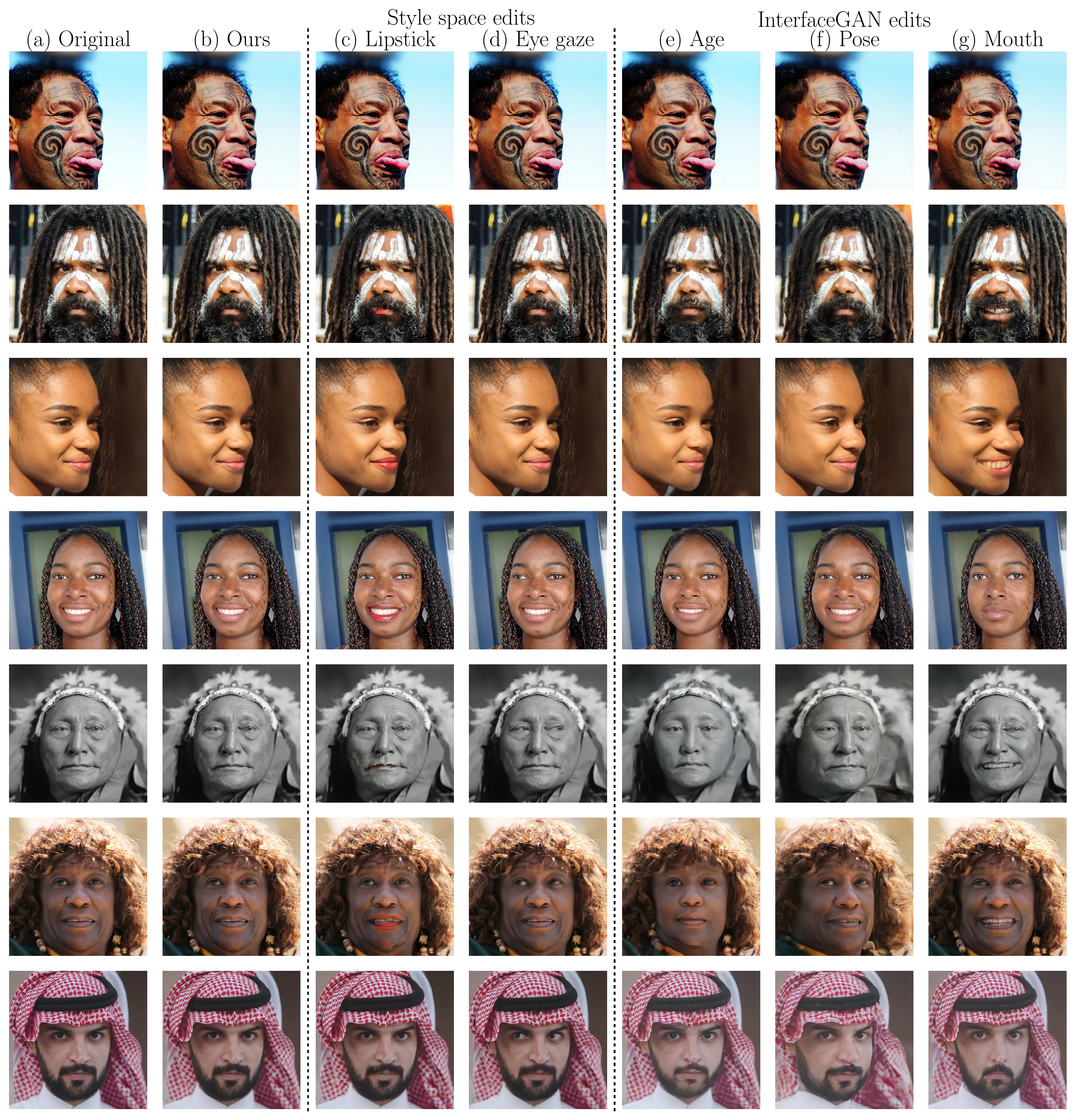}
  \caption{Qualitative results of our proposed Clone algorithm on under-represented sample images downloaded from the internet using a StyleGAN2 model trained on FFHQ.}
  \label{fig:6}
\end{figure*}

\newpage
\begin{figure*}[h]
  \centering
  \includegraphics[width=1.0\textwidth]{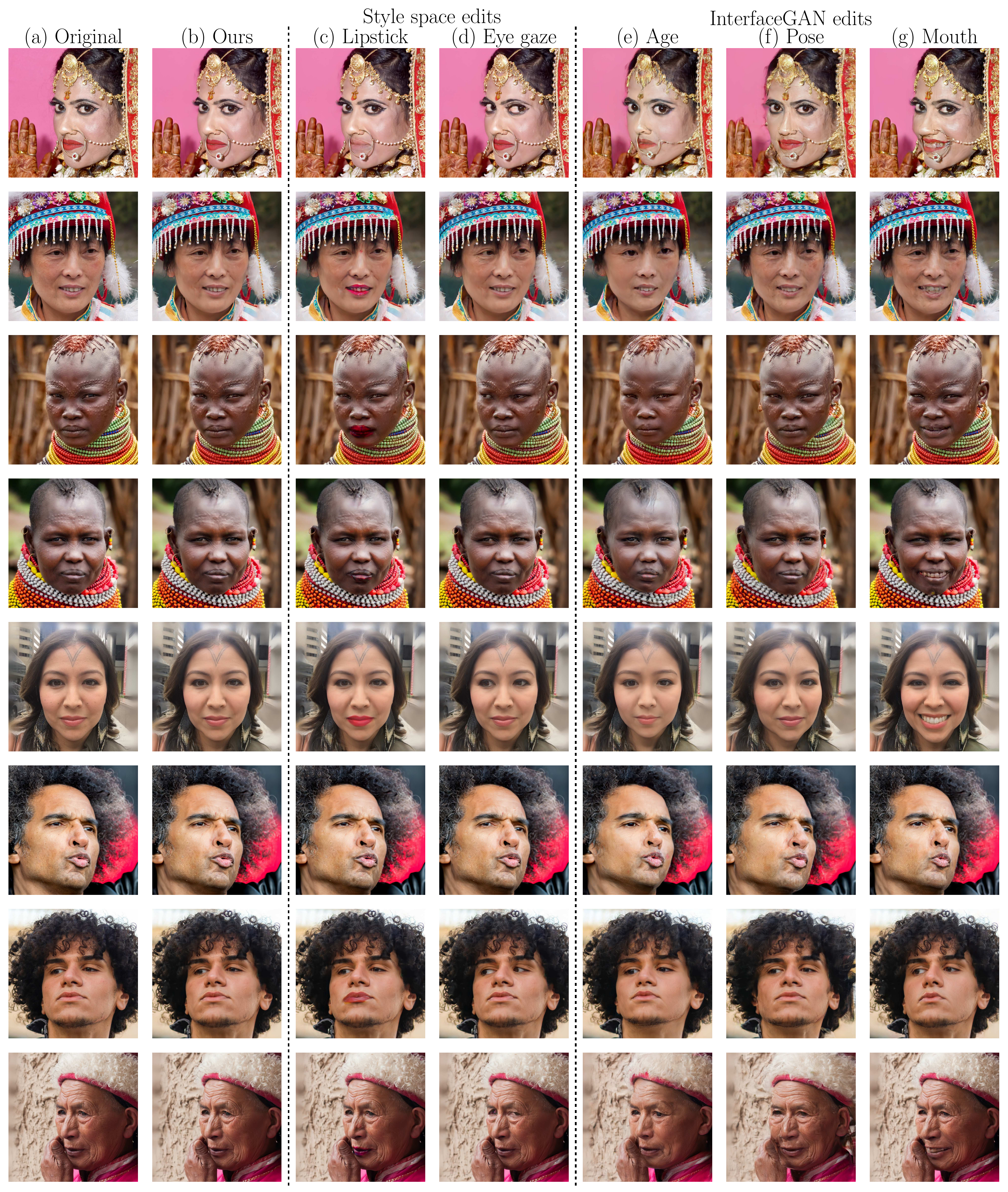}
  \caption{Qualitative results of our proposed Clone algorithm on under-represented sample images downloaded from the internet using a StyleGAN2 model trained on FFHQ.}
  \label{fig:7}
\end{figure*}

\newpage
\begin{figure*}[h!]
  \centering
  \includegraphics[width=1.0\textwidth]{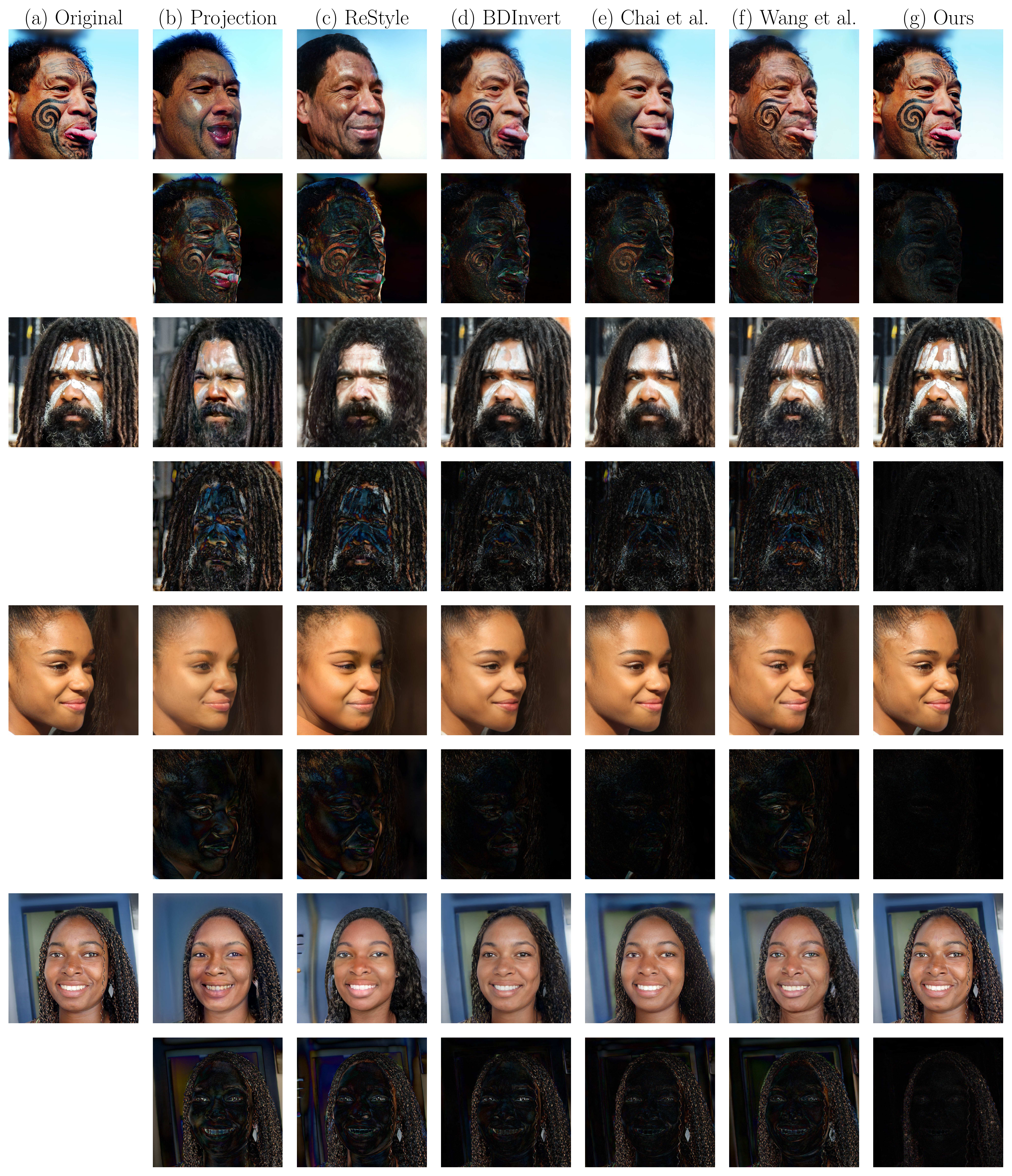}
  \caption{Synthetic image reconstructions as given by state-of-the-art GAN inversion methods and our proposed algorithm. Below each reconstruction, we show the pixel-to-pixel (2-norm) difference between the original photo and the synthesized image.}
  \label{fig:8}
\end{figure*}

\newpage
\begin{figure*}[h]
  \centering
  \includegraphics[width=1.0\textwidth]{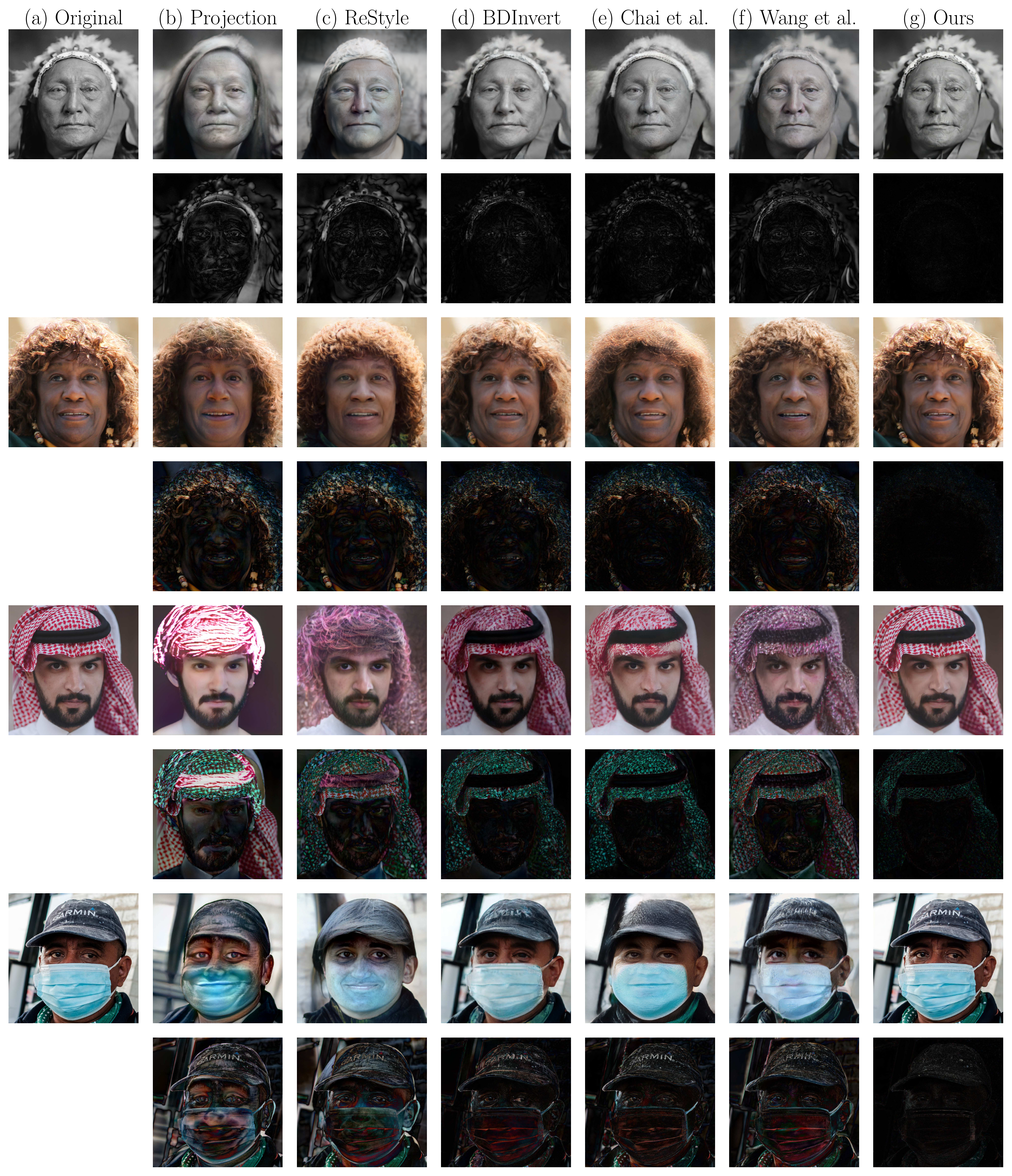}
  \caption{Synthetic image reconstructions as given by state-of-the-art GAN inversion methods and our proposed algorithm. Below each reconstruction, we show the pixel-to-pixel (2-norm) difference between the original photo and the synthesized image.}
  \label{fig:9}
\end{figure*}

\newpage
\begin{figure*}[h]
  \centering
  \includegraphics[width=1.0\textwidth]{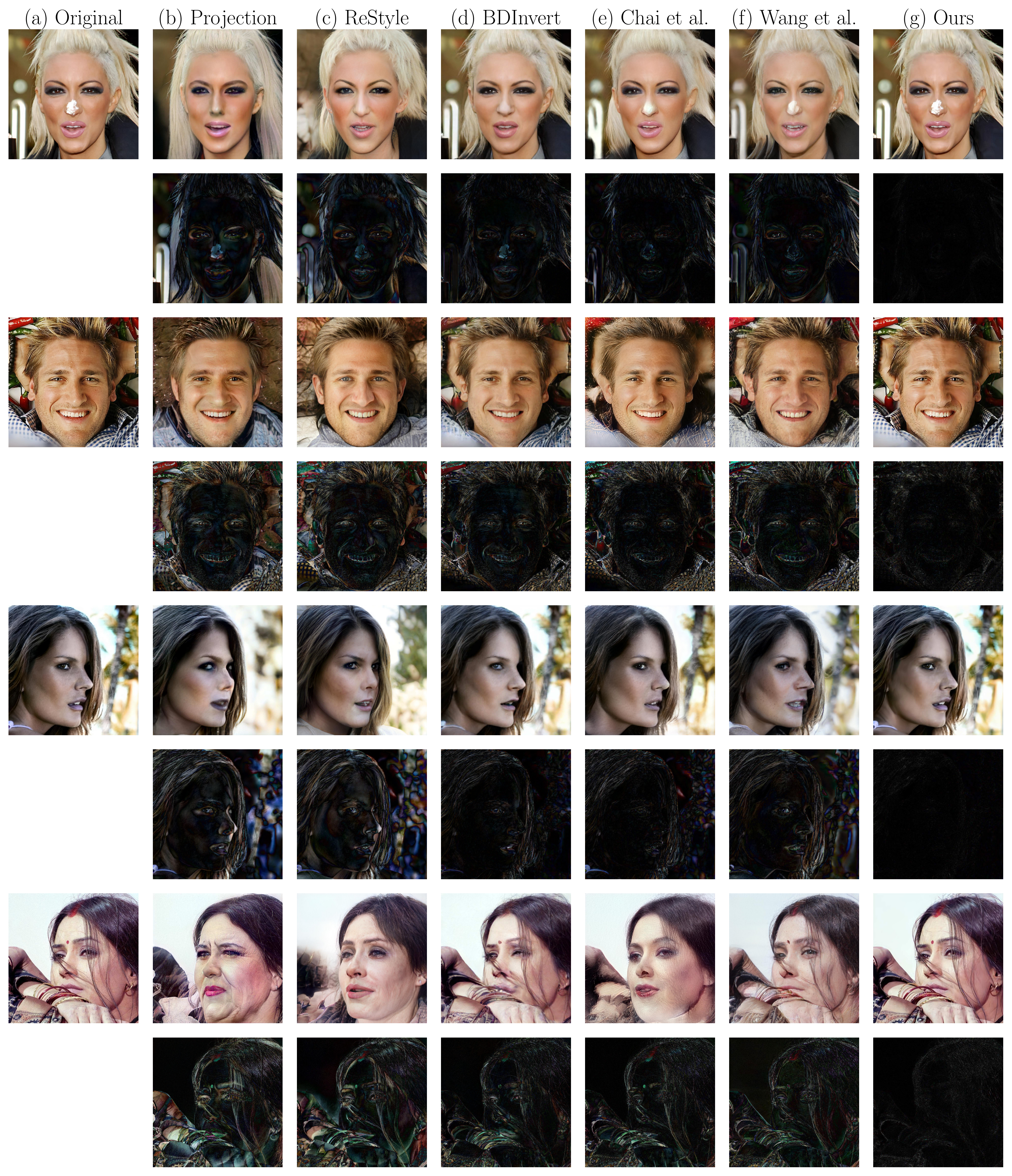}
  \caption{Synthetic image reconstructions as given by state-of-the-art GAN inversion methods and our proposed algorithm. Below each reconstruction, we show the pixel-to-pixel (2-norm) difference between the original photo and the synthesized image.}
  \label{fig:10}
\end{figure*}

\begin{figure*}[h!]
  \centering
  \includegraphics[width=1.0\textwidth]{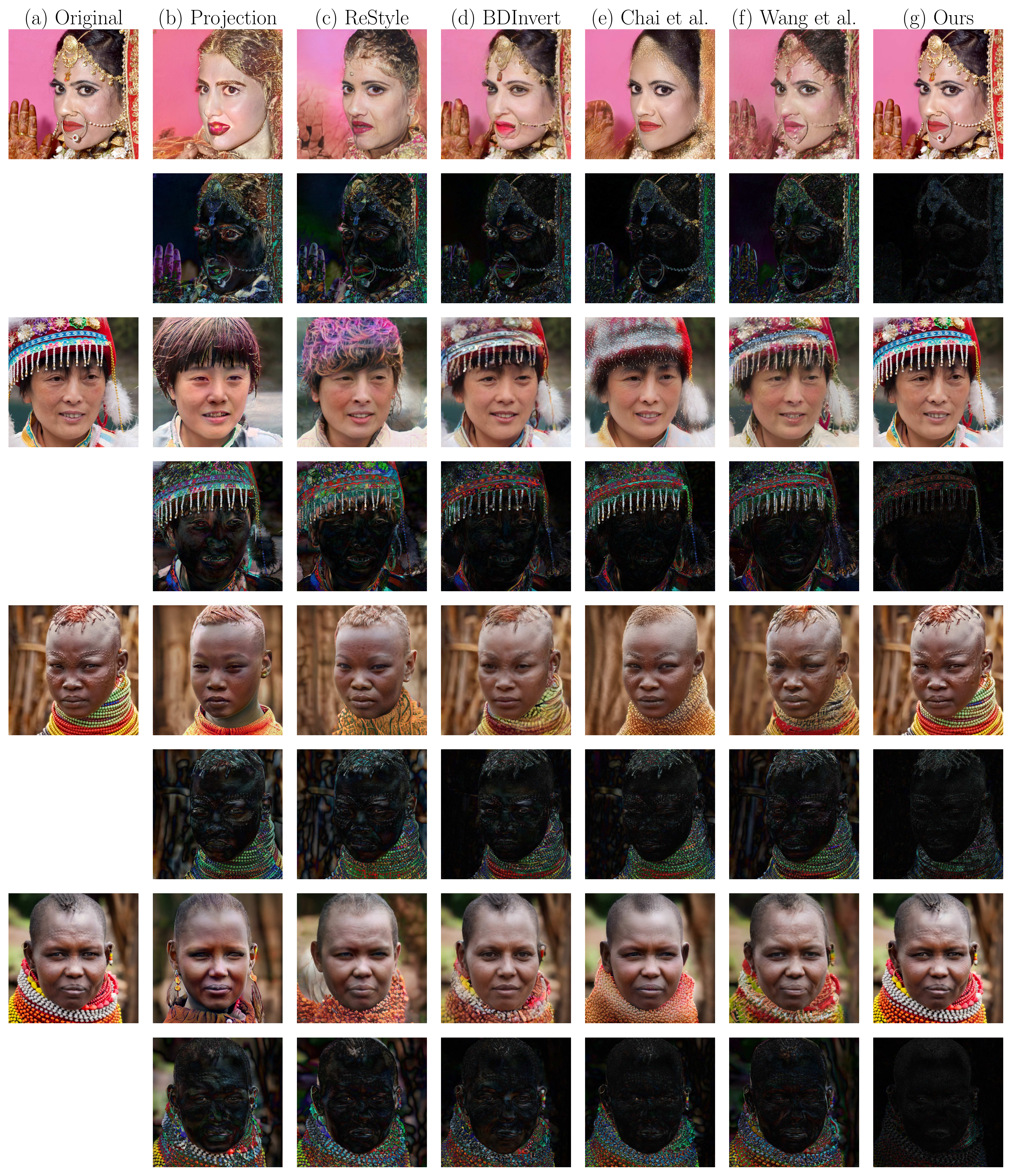}
  \caption{Synthetic image reconstructions as given by state-of-the-art GAN inversion methods and our proposed algorithm. Below each reconstruction, we show the pixel-to-pixel (2-norm) difference between the original photo and the synthesized image.}
  \label{fig:11}
\end{figure*}

\newpage
\begin{figure*}[h!]
  \centering
  \includegraphics[width=1.0\textwidth]{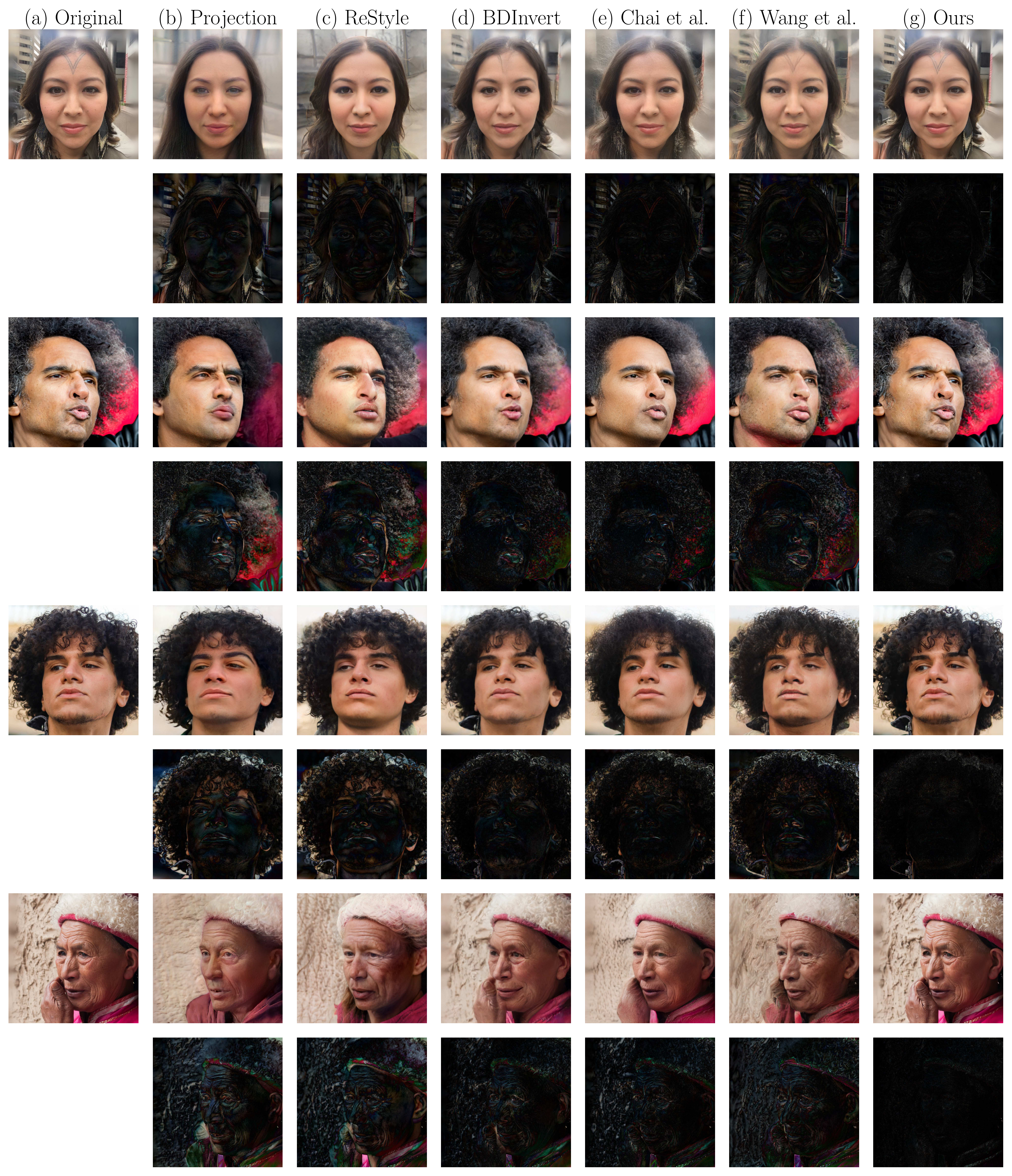}
  \caption{Synthetic image reconstructions as given by state-of-the-art GAN inversion methods and our proposed algorithm. Below each reconstruction, we show the pixel-to-pixel (2-norm) difference between the original photo and the synthesized image.}
  \label{fig:12}
\end{figure*}


\begin{figure*}[h]
  \centering
  \includegraphics[width=1.0\textwidth]{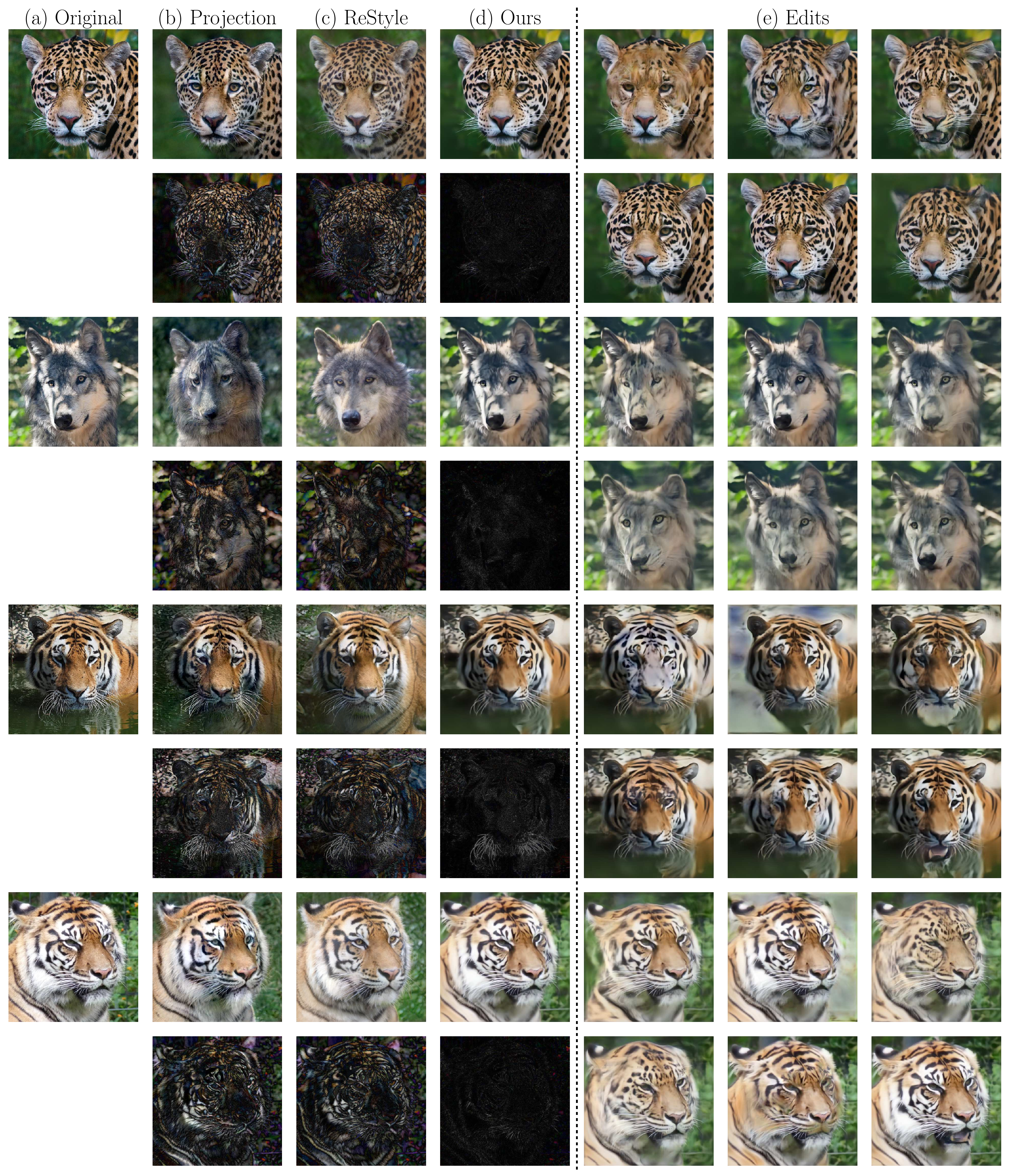}
  \caption{Additional comparative qualitative image synthesis and edits on images of animal faces. These images are from AFHQ-Wild. Below each reconstruction, we show the pixel-to-pixel (2-norm) difference between the original photo and the synthesized image.}
  \label{fig:13}
\end{figure*}

\newpage
\begin{figure*}[h]
  \centering
  \includegraphics[width=1.0\textwidth]{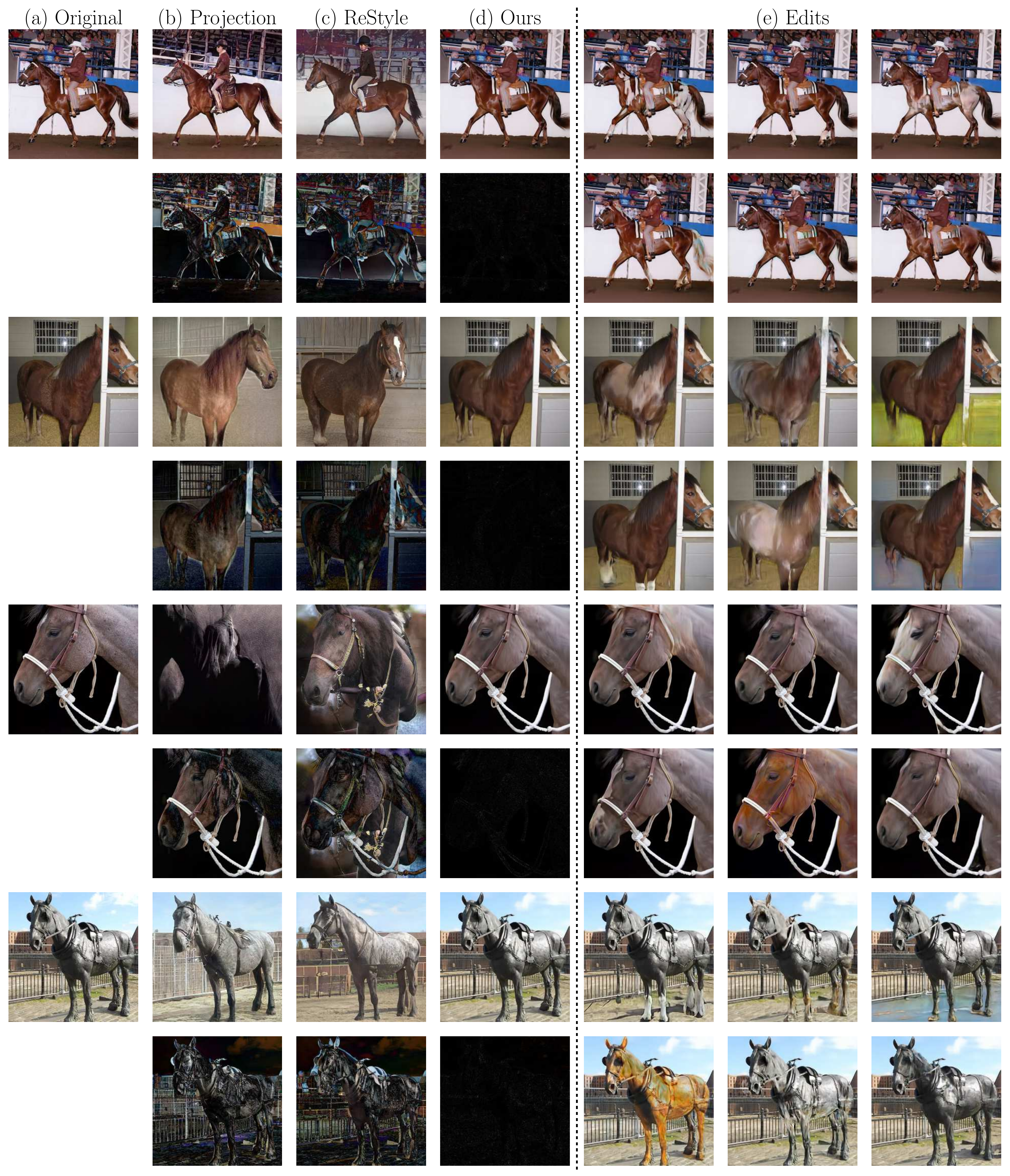}
  \caption{Additional comparative qualitative image synthesis and edits on images of horses. These images are from LSUN-Horses. Below each reconstruction, we show the pixle-to-pixel (2-norm) difference between the original photo and the synthesized image.}
  \label{fig:14}
\end{figure*}

\newpage
\begin{figure*}[h]
  \centering
  \includegraphics[width=1.0\textwidth]{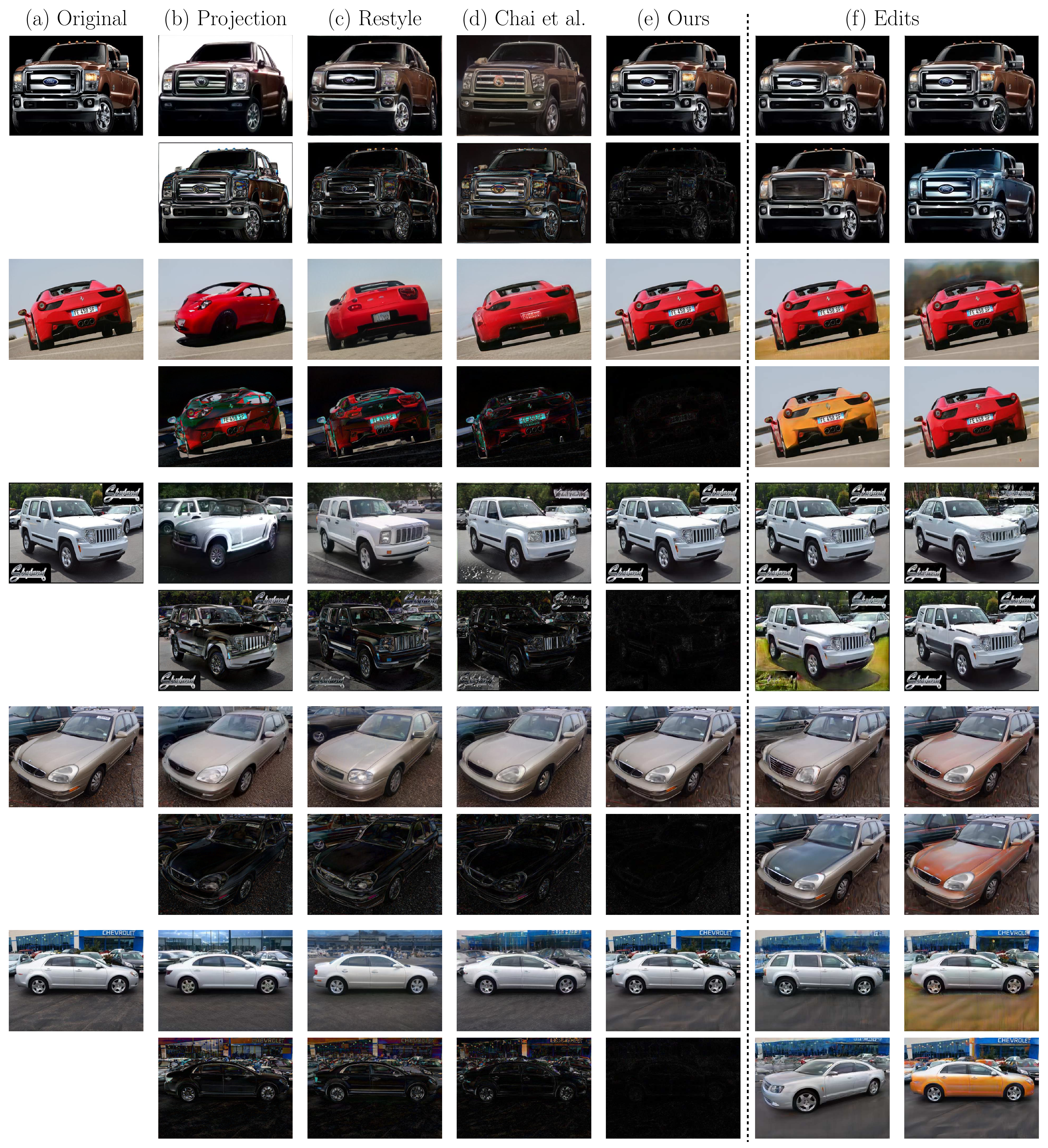}
  \caption{Additional comparative qualitative image synthesis and edits on images of cars. These images are from Stanford-Cars. Below each reconstruction, we show the pixle-to-pixel (2-norm) difference between the original photo and the synthesized image.}
  \label{fig:15}
\end{figure*}

\clearpage

\newpage
\section{Supplementary Documentation Overview}

This Supplementary File provides additional details of the proposed algorithm and experimental results. We also provide additional results and illustrate the use of the proposed approach on novel applications. \cref{sec:training-details} provides additional details on the training of the models used in the main paper. \cref{sec:tuning-heuristic} describes the heuristics used for hyper-parameter selection. \cref{sec:limitation} further details the assumptions and limitation of our algorithm. \cref{sec:social-impact} discusses ethical and potential societal impact of the work. \cref{fig:1,fig:2,fig:3,fig:4,fig:5,fig:6,fig:7,fig:8,fig:9,fig:10,fig:11,fig:12,fig:13,fig:14,fig:15} provides additional qualitative results for real photo inversions on human faces, cars, animal faces and horses.

\section{Model Training Details}\label{sec:training-details}

In the Faces experiment, we use a StyleGAN2 model pre-trained on FFHQ dataset. The quantitative evaluation is performed on a set of 500 CelebA-HQ images selected at random with no pre-processing. In our qualitative evaluation, we show inversions and editings on out-of-sample images selected from the internet. In these images, we first detect the face and estimate the facial landmarks using Dlib facial landmarks estimator \footnote{https://github.com/davisking/dlib}. The face is then scaled, translated and rotated to an upfront canonical position. Then, we apply a Gaussian blur around the image boundary to emphasize the face. The resolution of the images used in this experiment is $1024\times1024$ pixels.

In the Cars experiment, we use a StyleGAN2 model trained on LSUN-Cars. Our test is based on 500 images from the Stanford Cars database selected at random, with no pre-processing performed. The results provided by the ReStyle encoder are done at a $512\times384$ pixel resolution, while the generator is pre-trained to produce images at $512\times512$ pixels with top-bottom zero-padding. We follow the same practice which means that the $\mathcal{L}_\text{recon}$ is performed only on the $512\times384$ center crop while both $\mathcal{L}_\text{adv-local}$ and $\mathcal{L}_\text{global}$ are applied on the full $512\times512$-pixel image (with padding). 

In the Animal Faces experiment, we use a StyleGAN2 trained on the AFHQ-Wild training set and test on the full AFHQ-Wild test set (500 images), with no pre-processing. The image resolution in this experiment is $512\times512$ pixels.

In the Horses experiment, we use a StyleGAN2 model trained on the LSUN-Horses dataset and randomly sample 500 testing images from the full set. No pre-processing is performed. The image resolution in this experiment is $256\times256$ pixels.

\section{Hyper-Parameter Selection}\label{sec:tuning-heuristic}

As described in the main paper, we use different sets of hyper-parameter values for different experiments. A value of StyleGAN2 hyper-parameter $\gamma$ is also necessary as in the original StyleGAN2 training. \cref{tab:hyper-paramters} shows the parameter values used for each experiment shown in the main paper. 

\begin{table}[h!]
  \centering
  \begin{tabular}{@{}l c c c}
    \toprule
    Experiment 					& $p$ 	& $\lambda$ 	& $\gamma$ \\
    \midrule
    Faces (CelebA-HQ, FFHQ) 		& 1/4 	& 10			& 10 \\
    Cars (Stanford-Cars, LSUN-Cars) & 1/8 	& 10			& 5 \\
    AFHQ-wild 					& 1/32  	& 20 			& 15 \\
    LSUN-Horses 				& 1/16 	& 10			& 10 \\
    \bottomrule
  \end{tabular}
  \caption{Hyper-parameter values used in the experiments. $\gamma$ (gamma) is a StyleGAN2 specific parameter.}
  \label{tab:hyper-paramters}
\end{table}

The heuristic we used to select the parameters is: 
\begin{enumerate}
\item Initialize with $p=1/8$, $\lambda=10$ and the value of $\gamma$ equal to that in the pre-training model.
\item When both FID and reconstruction error are high, change GAN model specific hyper-parameters (e.g., gamma in StyleGAN2)
\item When FID is low but reconstruction error is high, increase $\lambda$.
\begin{itemize}
\item if this leads to increase FID, then decrease $p$.
\end{itemize}
\item When reconstruction error is low but FID is high, decrease $p$.
\begin{itemize}
\item if this leads to increase reconstruction error, then increase $\lambda$.
\end{itemize}
\end{enumerate}

\section{Assumptions and Limitation}\label{sec:limitation}

The proposed approach assumes that the pre-trained GAN model is able to generate photo-realistic images. If the GAN model is not able to generate photo-realistic images or has not been properly trained, the algorithm defined in this paper will not be able to find a synthetic clone.

Additionally, the almost perfect GAN inversion results shown above come at an additional small computational cost compared to previous methods. Since previous algorithms optimize $h(\cdot)$, their cost is associated to the number of iterations required to get good convergence. Optimization-based approaches generally require more iterations and, hence, are typically more expensive than encoder-based methods which require only one or a few iterations. In practice, this means optimization-based approaches may take up to a few minutes to solve the GAN inversion problem whereas encoder-based approaches may provide results in a few seconds of less.

The approach proposed in this paper, locally tweaks $G(\cdot)$. It needs to be noted that this is {\em not} the same as re-training the GAN model. This local tweak is successfully completed in just a few iterations, taking typically several seconds and up to a few minutes. In the worse cases, where our algorithm's solution ${\bf x}^*$ is significantly far from $\hat{\bf x}$, the {\em Clone} algorithm may take several minutes. This may limit the use of the proposed technique in applications that require close to real-time results.

\section{Potential Societal Impact}\label{sec:social-impact}

This paper poses a similar societal impact as other GAN inversion techniques or realistic image editing methods. 

On the one hand, the near perfect inversion and editing can be used for editing memorable photos, helping people re-experience the moments they treasured. The method can also provide a chance to correct unsatisfactory poses and expressions after a photo has been captured, especially when the photo cannot be easily re-taken. We believe these new editing opportunities provided by our algorithm can be used to save time and making valuable memory more vidid and lively. There are also a number of potential applications in film and e-commerce that were not possible with previous GAN inversion algorithms due the lack of almost-perfect reconstructions. Moreover, as described in the main paper, our method can reconstruct photos even if the people in the photo are under-represented in the training set. 

On the other hand, the method can be used to manipulate photos to spread misinformation by changing eye gaze, facial expressions and other semantic attributes. The negative aspect of photo editing is not a new concern, as realistic editing can be already achieved using ``Photoshoping'' and other computer graphics, computer vision and machine learning techniques. Thus, we think it is important to develop algorithms that can tell these edited images apart. 
AI regulations may also be warranted. 

\section{Image Attribution}\label{sec:license-attribution}

This paper uses and edits the images with Creative Commons license described in \cref{tab:license-info}

\begin{table*}[h!]
  \centering
  \begin{tabular}{m{11cm} c c}
    \toprule
    URL 									& Authors 			& License \\
    \midrule
    \url{https://commons.wikimedia.org/wiki/File:M%C3%A9lanie_de_Jesus_dos_Santos_-_Glasgow_2018_-_01_cropped.jpg}				& TwoWings 			& CC-BY-SA-4.0	\\
    \url{https://commons.wikimedia.org/wiki/File:Alice_in_Chains_-_2019158181605_2019-06-07_Rock_am_Ring_-_1163_-_B70I8463.jpg}	& Sven Mandel 		& CC-BY-SA-4.0	\\
    \url{https://commons.wikimedia.org/wiki/File:Bahjat_Artist_.jpg}															& Teambahjat  			&CC-BY-SA-4.0	\\
    \url{https://commons.wikimedia.org/wiki/File:Young_Woman_%28Adriana%29_at_Tourist_Office_-_Cachoeira_-_Bahia_-_Brazil.JPG}		& Adam Jones\footnote{adamjones.freeservers.com}			& CC-BY-SA-3.0			\\
     \url{https://commons.wikimedia.org/wiki/File:Maori_Man_%28Imagicity_1034%29.jpg}											& Graham Crumb \footnote{http://www.imagicity.com/}			& CC-BY-SA-3.0			\\
     \url{https://commons.wikimedia.org/wiki/File:Alethea_Arnaquq-Baril_in_Toronto.jpg}											& Aarnaquq			& CC-BY-SA-4.0			\\
     \url{https://live.staticflickr.com/65535/49542075471_9fde3d42a5_o_d.jpg}											& Rod Waddington			& CC-BY-SA-2.0			\\
     \url{https://live.staticflickr.com/65535/49533727621_0e254349bb_o_d.jpg}											& Rod Waddington			& CC-BY-SA-2.0			\\
     \url{https://www.maxpixel.net/Warrior-Man-Strong-2152851}														& Max Pixel 		& CC0 Public Domain			\\
    \bottomrule
  \end{tabular}
  \caption{Image license information}
  \label{tab:license-info}
\end{table*}

{\small
\bibliographystyle{ieee_fullname}
\bibliography{references}
}